\documentclass[10pt,twocolumn,letterpaper]{article}

\usepackage{cvpr}
\usepackage{times}
\usepackage{epsfig}
\usepackage{graphicx}
\usepackage{amsmath}
\usepackage{amssymb}
\usepackage{graphicx,multirow}
\usepackage{subfig}
\usepackage{url}
\usepackage{caption}
\usepackage{overpic}
\usepackage{wrapfig}

\usepackage{amsthm}
\usepackage{algorithm}
\usepackage{siunitx}
\usepackage{algorithmicx}
\usepackage{algpseudocode}

\usepackage{parskip}
\usepackage{caption}
\usepackage{mdwlist}

\newcommand{\R}{\mathbb{R}}
\let\bs=\boldsymbol

\def \diag {\mathrm{diag}}

\def \train {\textup{train}}

\def \saliency {\textup{\saliency}}

\def \path {\textup{path}}

\def \path {\mathit{path}}

\makeatother

\theoremstyle{plain}

 \theoremstyle{definition}
\numberwithin{theorem}{section}
\numberwithin{lem}{section}

\theoremstyle{remark}

  \let\set=\mathcal 
\newcommand{\qixing}[1]{\textcolor{red}{[#1]}}

 \global\long\def\diag{\mathrm{diag}}

 \global\long\def\R{\mathbb{R}}

\usepackage[pagebackref=true,breaklinks=true,colorlinks,bookmarks=false]{hyperref}

\cvprfinalcopy 

\def\cvprPaperID{3439} 

\begin{document}

\title{Extreme Relative Pose Estimation for RGB-D Scans via Scene Completion}

\author{Zhenpei Yang\\
UT Austin
\and
Jeffrey Z.Pan\\
Phillips Academy Andover
\and
Linjie Luo\\
Snap Research
\and
Xiaowei Zhou\\
Zhejiang University
\and
Kristen Grauman\\
Facebook AI Research\thanks{On leave from University of Texas at Austin (grauman@cs.utexas.edu).}
\and
Qixing Huang\\
UT Austin
}

\maketitle

\begin{abstract}
Estimating the relative rigid pose between two RGB-D scans of the same underlying environment is a fundamental problem in computer vision, robotics, and computer graphics. Most existing approaches allow only limited maximum relative pose changes since they require considerable overlap between the input scans. We introduce a novel deep neural network that extends the scope to extreme relative poses, with little or even no overlap between the input scans. The key idea is to infer more complete scene information about the underlying environment and match on the completed scans. In particular, instead of only performing scene completion from each individual scan, our approach alternates between relative pose estimation and scene completion. This allows us to perform scene completion by utilizing information from both input scans at late iterations, resulting in better results for both scene completion and relative pose estimation. Experimental results on benchmark datasets show that our approach leads to considerable improvements over state-of-the-art approaches for relative pose estimation. In particular, our approach provides encouraging relative pose estimates even between non-overlapping scans. 
\end{abstract}

\section{Introduction}
\label{Section:Introduction}

Estimating the relative rigid pose between a pair of RGB-D scans is a fundamental problem in computer vision, robotics, and computer graphics with applications to systems such as 3D reconstruction~\cite{DBLP:journals/cgf/ZollhoferSGTNKK18}, structure-from-motion~\cite{Snavely:2006:PTE}, and simultaneous localization and mapping (SLAM)~\cite{conf/iros/SturmEEBC12}. Most existing approaches~\cite{Gelfand:2005:RGR,Huang:2006:RFO,Aiger:2008:CSR,CGF:CGF12446,DBLP:conf/eccv/ZhouPK16} follow a three-step paradigm (c.f.~\cite{DBLP:journals/cgf/ZollhoferSGTNKK18}): feature extraction, feature matching, and rigid transform fitting with the most consistent feature correspondences. However, this paradigm requires the input RGB-D scans to have considerable overlap, in order to establish sufficient feature correspondences for matching. For input scans of extreme relative poses with little or even \emph{no} overlap, this paradigm falls short since there are very few or no features to be found in the overlapping regions. Nevertheless, such problem settings with minimal overlap are common in many applications such as solving jigsaw puzzles~\cite{Cho_TPAMI2010}, early detection of loop closure for SLAM~\cite{Henry:2012:RMU}, and reconstruction from minimal observations, e.g., a few snapshots of an indoor environment~\cite{DBLP:conf/cvpr/LiuSKUF15}. 

\begin{figure}
\centering
\includegraphics[width=1.0\columnwidth,trim=510 290 500 290,clip]{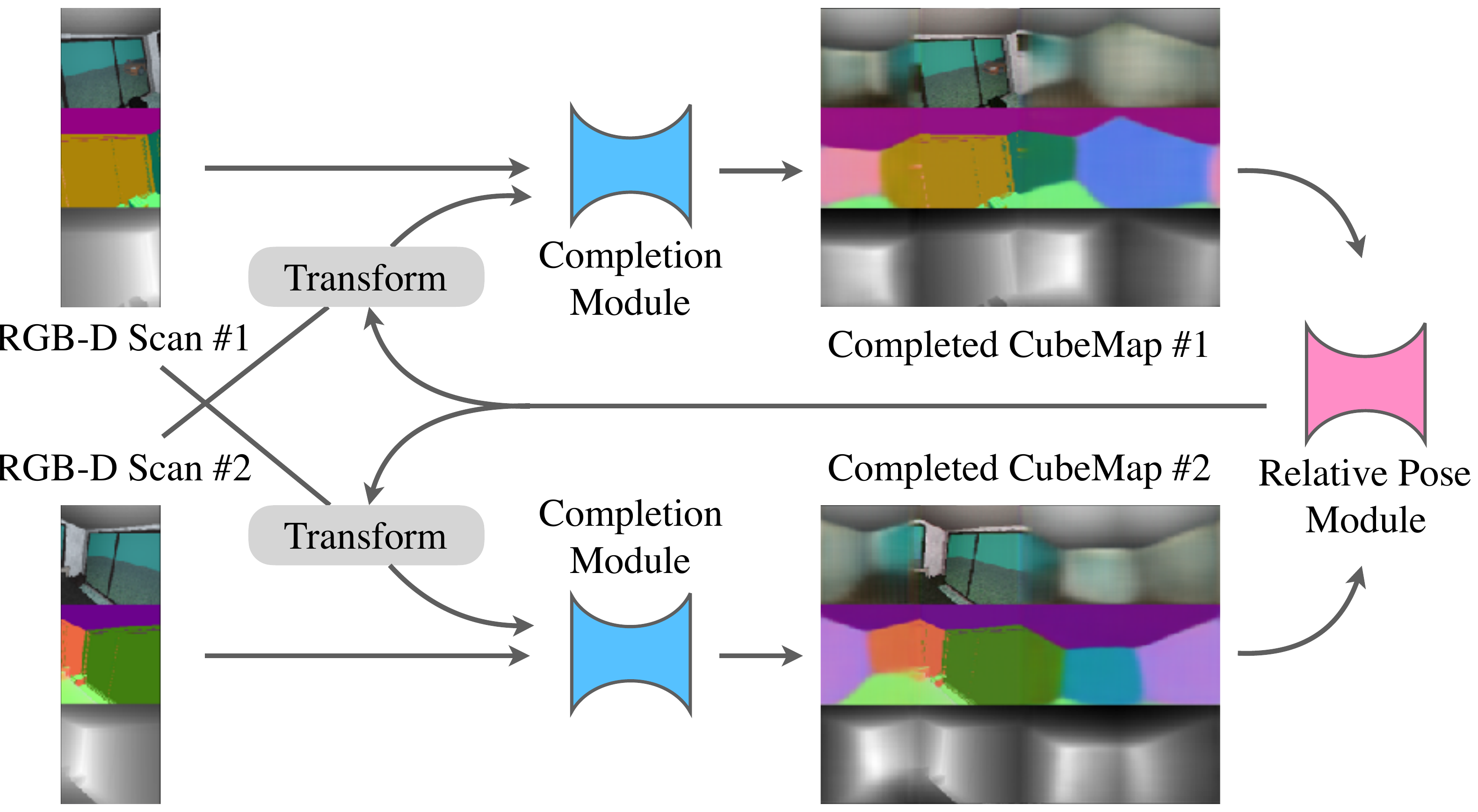}
\caption{Illustration of the work-flow of our approach. We align two RGB-D scans by alternating between scene completion (completion module) and pose estimation (relative pose module).}
\label{Figure:Work:Flow}
\vspace{-0.15in}
\end{figure}

While the conventional paradigm breaks down in this setting, we hypothesize that solutions are possible using the prior knowledge for typical scene structure and object shapes. Intuitively, when humans are asked to perform pose estimation for non-overlapping inputs, they utilize the prior knowledge of the underlying geometry. For example, we can complete a human model from two non-overlapping scans of both the front and the back of a person; we can also tell the relative pose of two non-overlapping indoor scans by knowing that the layout of the room satisfies the Manhattan world assumption~\cite{Coughlan:1999:MWC}. This suggests that when direct matching of non-overlapping scans is impossible, we seek to match them by first performing scene completions and then matching completed scans for their relative pose.

Inspired from the iterative procedure for simultaneous reconstruction and registration~\cite{DBLP:conf/icra/HuangA10}, we propose to alternate between scene completion and relative pose estimation so that we can leverage signals from both input scans to achieve better completion results.  
Specifically, we introduce a neural network that takes a pair of RGB-D scans with little overlap as input and outputs the relative pose between them.
Key to our approach are internal modules that infer the \emph{completion} of each input scan, allowing even widely separated scans to be iteratively registered with the proper relative pose via a recurrent module. As highlighted in Figure~\ref{Figure:Work:Flow}, our network first performs single-scan completion under a rich representation that combines depth, normal, and semantic descriptors. This is followed by a pair-wise matching module, which takes the current completions as input and outputs the current relative pose. 

In particular, to address the issue of imperfect predictions, we introduce a novel pairwise matching approach that seamlessly integrates two popular pairwise matching approaches: spectral matching~\cite{Leordeanu:2005:STC,Huang:2008:NRU} and robust fitting~\cite{Bouaziz:2013:SIC}. Given the current relative pose, our network performs bi-scan completion, which takes as input the representation of one scan and transformed representation of the other scan (using the current relative pose estimate), and outputs a updated scan completion in the view of the first scan. The pair-wise matching module and the bi-scan completion module are alternated, as  reflected by the recurrent nature of our network design.   
Note that compared to existing deep learning based pairwise matching approaches~\cite{DBLP:journals/corr/LongZD14,DBLP:journals/corr/FischerDIHHGSCB15}, which combine feature extraction towers and a matching module, the novelty of our approach is three-fold:

\begin{enumerate*}
  \item explicitly supervising the relative pose network via completions of the underlying scene under a novel representation that combines geometry and semantics.
  \item a novel pairwise matching method under this representation.
  \item an iterative procedure that alternates between scene completion and pairwise matching.
\end{enumerate*}

We evaluate our approach on three benchmark datasets, namely, SUNCG~\cite{song2016ssc}, Matterport~\cite{DBLP:journals/corr/abs-1709-06158}, and ScanNet~\cite{DBLP:journals/corr/DaiCSHFN17}.
Experimental results show that our approach is significantly better than state-of-the-art relative pose estimation techniques. For example, our approach reduces the mean rotation errors of state-of-the-art approaches from \ang{36.6}, \ang{42.0}, and \ang{51.4} on SUNCG, Matterport, and ScanNet, respectively, to \ang{12.0}, \ang{9.0}, and \ang{30.2}, respectively, on scans with overlap ratios greater than 10\%. Moreover, our approach generates encouraging results for non-overlapping scans. The mean rotation errors of our approach for these scans are \ang{79.9}, \ang{87.9}, and \ang{81.8}, respectively. In contrast, the expected error of a random rotation is around \ang{126.3}. 

Code is publicly available at \url{https://github.com/zhenpeiyang/RelativePose}.

\section{Related Work}
\label{Section:Related:Works}

\noindent\textbf{Non-deep learning techniques.} Pairwise object matching has been studied extensively in the literature, and it is beyond the scope of this paper to present a comprehensive overview. We refer to~\cite{10.1111:j.1467-8659.2011.01884.x,Tangelder:2008:SCB,Li:2014:CMO} for surveys on this topic and to~\cite{CGF:CGF12446} for recent advances. Regarding the specific task of relative pose estimation from RGB-D scans, popular methods~\cite{Gelfand:2005:RGR,Huang:2006:RFO,Aiger:2008:CSR,CGF:CGF12446} follow a three-step procedure. The first step extracts features from each scan. The second step establishes correspondences for the extracted features, and the third step fits a rigid transform to a subset of consistent feature correspondences. Besides the fact that the performance of these techniques heavily relies on setting suitable parameters for each component, they also require that the two input scans possess sufficient overlapping features to match.   

\noindent\textbf{Deep learning techniques.} Thanks to the popularity of deep neural networks, recent work explores
deep neural networks for the task of relative pose estimation (or pairwise matching in general) ~\cite{DBLP:journals/corr/FischerDIHHGSCB15,DBLP:journals/corr/IlgMSKDB16,ummenhofer2017demon,zhou2017unsupervised,melekhov2017relative}. These approaches follow the standard pipeline of object matching, but they utilize a neural network module for each component. Specifically, feature extraction is generally done using a feed-forward module, while estimating correspondences and computing a rigid transform 

are achieved using a correlation module. With proper pre-training, these methods exhibit better performance than their non-deep learning counterparts. However, they still require that the inputs possess a sufficient overlap so that the correlation module can identify common features for relative pose estimation.  

A couple of recent works propose 
recurrent procedures for object matching. In~\cite{ranftl2018deep}, the authors present a recurrent procedure to compute weighted correspondences for estimating the fundamental matrix between two images. In~\cite{DBLP:conf/nips/KimLJMS18}, the authors use recurrent networks to progressively compute dense correspondences between two images. The network design is motivated from the procedure of non-rigid image registration between a pair of images. Our approach is conceptually relevant to these approaches. However, the underlying principle for the recurrent approach is different. In particular, our approach performs scan completions, from which we compute the relative pose.

\noindent\textbf{Optimization techniques for pairwise matching.} Existing feature-based pairwise matching techniques fall into two categories. The first category of methods is based on MAP inference~\cite{Leordeanu:2005:STC,Huang:2008:NRU,Chen:2015:RNR}, where feature matches and pairwise feature consistency are integrated as unary and pairwise potential functions. A popular relaxation of MAP inference is spectral relaxation~\cite{Leordeanu:2005:STC,Huang:2008:NRU}. The second category of methods is based on fitting a rigid transformation to a set of feature correspondences~\cite{Horn87closed-formsolution}. In particular, state-of-the-art approaches~\cite{Eggert:1997:ERB,10.1111:j.1467-8659.2011.01884.x,DBLP:conf/eccv/ZhouPK16} usually utilize robust norms to handle incorrect feature correspondences. In this paper, we introduce the first approach that optimizes a single objective function to simultaneously perform spectral matching and robust fitting for relative pose estimation. 
\begin{figure*}
\includegraphics[width=\textwidth,trim=30 360 30 360,clip]{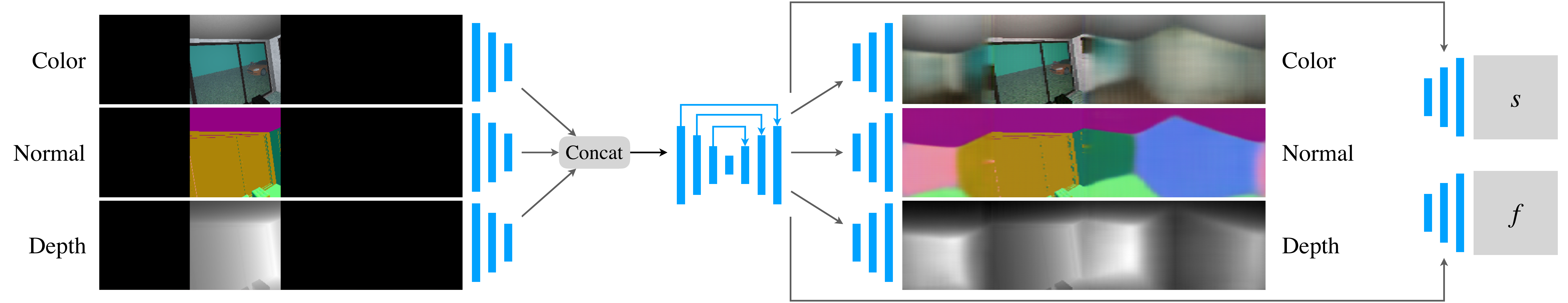}

\caption{\small{Network design of the completion module. Given the partially observed color, depth, normal, our network complete cube-map representation of color, depth, normal, semantic, as well as a feature map. Please refer to Sec.~\ref{Section:Scan:Completion} for details.}}
\label{Figure:Single:Scan:Completion:Module}
\vspace{-0.2in}
\end{figure*}

\noindent\textbf{Scene completion.} Our approach is also motivated from recent advances on inferring complete environments from partial observations~\cite{DBLP:journals/corr/PathakKDDE16,song2016im2pano3d,DBLP:journals/corr/abs-1709-00505,Zou_2018_CVPR}. However, our approach differs from these approaches in two ways. First, in contrast to returning the completion as the final output~\cite{song2016im2pano3d,Zou_2018_CVPR} or utilizing it for learning feature representations~\cite{DBLP:journals/corr/PathakKDDE16,DBLP:journals/corr/abs-1709-00505}, our approach treats completions as an intermediate representation for relative pose estimation. From the representation perspective, our approach predicts color,depth,normal,semantic,and feature vector using a single network. Experimentally, this leads to better results than performing completion using the RGB-D representation first and then extracting necessary features from the completions.

\section{Approach}
\label{Section:Network:Design}

We begin with presenting an approach overview in Section~\ref{Subsection:Approach:Overview}. Section~\ref{Section:Feature:Representation} to Section~\ref{Section:Relative:Pose:Module} elaborate the network design. Section~\ref{Section:Network:Training} discusses the training procedure. 

\subsection{Approach Overview}
\label{Subsection:Approach:Overview}
The relative pose estimation problem studied in this paper considers two RGB-D scans $I_i\in \R^{h\times w\times 4}, 1\leq i \leq 2$ of the same environment as input ($h$,$w$=160 in this paper). We assume that the intrinsic camera parameters are given so that we can extract the 3D position of each pixel in the local coordinate system of each $I_i$. The output of relative pose estimation is a rigid transformation $T = (R,\bs{t})\in \R^{3\times 4}$ that recovers the relative pose between $I_1$ and $I_2$. Note that we do not assume $I_1$ and $I_2$ overlap.   

Our approach is inspired from simultaneous registration and reconstruction (or SRAR)~\cite{DBLP:conf/icra/HuangA10}, which takes multiple depth scans of the same environment as input and outputs both a 3D reconstruction of the underlying environment (expressed in a world coordinate system) and optimized scan poses (from which we can compute relative poses). The optimization procedure of SRAR alternates between fixing the scan poses to reconstruct the underlying environment and optimizing scan poses using the current 3D reconstruction. The key advantage of SRAR is that pose optimization can leverage a complete reconstruction of the underlying environment and thus it mitigates the issue of non-overlap.

However, directly applying SRAR to relative pose estimation for 3D scenes is challenging, as unlike 3D objects~\cite{3dgan,DBLP:conf/cvpr/QiSNDYG16,DBLP:conf/eccv/ChoyXGCS16,NIPS2016_6206}, it is difficult to specify a world coordinate system for 3D scenes. To address this issue, we modify SRAR by maintaining two copies $S_1$ and $S_2$ of the complete underlying environment, where $S_i$ is expressed in the local coordinate system of $I_i$ (We will discuss the precise representation of $S_i$ later). Conceptually, our approach reconstructs each $S_i$ by combining the signals in both $I_1$ and $I_2$. When performing relative pose estimation, our approach employs $S_1$ and $S_2$, which addresses the issue of non-overlap.  

As illustrated in Figure~\ref{Figure:Work:Flow}, the proposed network for our approach combines a scan completion module and a pairwise matching module. 
To provide sufficient signals for pairwise matching, we define the feature representation $\overline{X}, X\in \{I_1,I_2,S_1,S_2\}$ by concatenating color, depth, normal, semantic label, and descriptors. Here $\overline{S}_i$ utilizes a reduced cube-map representation~\cite{song2016im2pano3d}, where each face of $\overline{S}_i$ shares the same representation as $\overline{I}_i$. 

Experimentally, we found this approach gives far better results than performing scan completion under the RGB-D representation first and then computing the feature representation. The pairwise matching module takes current $\overline{S}_1$ and $\overline{S}_2$ as input and outputs the current relative pose $T$. The completion module updates each scan completion using the transformed scans, e.g., $\overline{S}_1$ utilizes $\overline{I}_1$ and transformed $\overline{I}_2$ in the local coordinate system of $I_1$. We alternate between applying the pairwise matching module and the bi-scan completion module. In our implementation, we use $3$ recurrent steps. Next we elaborate on the details of our approach.

\subsection{Feature Representation}
\label{Section:Feature:Representation}
Motivated by the particular design of our pairwise matching module, we define the feature representation of an RGB-D scan $I$ as $\overline{I} = (\bs{c},\bs{d}, \bs{n}, \bs{s},f)$. Here  $\bs{c}\in \R^{h\times w\times 3}$, $\bs{d}\in \R^{h\times w\times 1}$, $\bs{n}\in \R^{h\times w\times 3}$, $\bs{s}\in \R^{h\times w\times n_c}$(we use $n_c$=15 for SUNCG, $n_c$=21 for Matterport/ScanNet), $\bs{f}\in \R^{h\times w\times k}$($k$=32 in this paper), specify color, depth, normal, semantic class, and a learned descriptor, respectively. 
The color,depth,normal,semantic class are obtained using the densely labeled reconstructed model for all datasets.
 %

\subsection{Scan Completion Modules}
\label{Section:Scan:Completion}
The scan completion module takes in a source scan, a target scan transformed by current estimate $T$, and output the complete feature representation $\overline{S}_i$.
We encode $\overline{S}_i$ using a reduced cube-map representation~\cite{song2016im2pano3d}, which consists of four faces (excluding the floor and the ceiling). Each face of $\overline{S}_i$ shares the same feature representation as $\overline{I}_i$. For convenience, we always write $\overline{S}_i$ in the tensor form as $\overline{S}_i = (\overline{S}_{i,1},\overline{S}_{i,2},\overline{S}_{i,3},\overline{S}_{i,4})\in \R^{h\times w\times 4(k+n_c+7)}$. Following the convention~\cite{song2016im2pano3d,DBLP:journals/corr/PathakKDDE16}, we formulate the input to both scan completion modules using a similar tensor form $\hat{I}_i = (\hat{I}_{i,1},\hat{I}_{i,2},\hat{I}_{i,3},\hat{I}_{i,4})\in \R^{h\times w\times 4(k+n_c+8)}$, where the last channel is a mask that indicates the presence of data. As illustrated in Figure~\ref{Figure:Single:Scan:Completion:Module}~(Left), we always place $\overline{I}_{i}$ in $\hat{I}_{i,2}$. This means $\hat{I}_{i,1}$, $\hat{I}_{i,3}$, and $\hat{I}_{i,4}$ are left blank. 

We adapt a convolution-deconvolution structure for our scan completion network, denoted $g_{\phi}$. As shown in Figure~\ref{Figure:Single:Scan:Completion:Module}, we use separate layers to extract information from color, depth, and normal input, and concatenate the resulting feature maps. Note that we stack the source and transformed target scan in each of the color, normal, depth components to provide the network more information. Only source scan is shown for simplicity.
Since designing the completion network is not the major focus of this paper, we leave the technical details to supplementary material. 


\subsection{Relative Pose Module}
\label{Section:Relative:Pose:Module}

We proceed to describe the proposed relative pose module denoted as $h_{\gamma}(\overline{S}_1,\overline{S}_2)\rightarrow (R,\bs{t})$. 
We first detect SIFT keypoints on observed region, and further extracts the top matches of the keypoints on the other complete scan to form the final point set $\set{Q}_i$.
With $\set{Q}_1$ and $\set{Q}_2$ we denote the resulting points. Our goal is to simultaneously extract a subset of correspondences from $\set{C} = \set{Q}_1\times \set{Q}_2$ and fit $(R,\bs{t})$ to these selected correspondences. For efficiency, we remove a correspondence $c =(q_1,q_2)$ from $\set{C}$ whenever $\exp(-\|\bs{f}(q_1) -\bs{f}(q_2)\|^2/2/\gamma_1^2) \leq 10^{-2}$.


The technical challenge of extracting correct correspondences is that due to imperfect scan completions, many correspondences with similar descriptors are still outliers. We address this challenge by combining spectral matching~\cite{Leordeanu:2005:STC} and robust fitting~\cite{Bouaziz:2013:SIC}, which are two complementary pairwise matching methods. Specifically, let $x_{c}\in \{0,1\}, \forall c \in \set{C}$ be latent indicators. We compute $(R,\bs{t})$ by solving
\begin{align}
\underset{\{x_c\},R,\bs{t}}{\textup{maximize}} & \ \sum\limits_{c,c'\in \set{C}} w_{\gamma}(c,c')x_{c}x_{c'}\big(\delta - r_{(R,\bs{t})}(c) - r_{(R,\bs{t})}(c')\big) \nonumber \\
\textup{subject to} & \ \sum\limits_{c\in \set{C}}x_{c}^2 = 1
\label{Eq:Objective:Function}
\end{align}
As we will define next, $w(c,c')$ is a consistency score associated with the correspondence pair $(c,c')$, and $r_{(R,\bs{t})}(c)$ is a robust regression loss between $(R,\bs{t})$ and $c$. 
$\delta$ is set to 50 in our experiments. 
Intuitively, (\ref{Eq:Objective:Function}) seeks to extract a subset of correspondences that have large pairwise consistency scores and can be fit well by the rigid transformation. 

We define $w(c,c')$, where $c = (q_1,q_2)$ and $c' = (q'_1,q'_2)$, by combining five consistency measures:
\begin{align}
\Delta_1^2(c,c') := &\|\bs{f}(q_1) -\bs{f}(q_2)\|^2+\|\bs{f}(q'_1) -\bs{d}(q'_2)\|^2\nonumber \\ 
\Delta_2(c,c') := &\|\bs{p}(q_1)  -\bs{p}(q'_1)\|-\|\bs{p}(q_2) -\bs{p}(q'_2)\| \nonumber \\
\Delta_3(c,c') := &\angle (\bs{n}(q_1),\bs{n}(q'_1))-\angle(\bs{n}(q_2),\bs{n}(q'_2)) \nonumber \\
\Delta_4(c,c') := &\angle (\bs{n}(q_1),\bs{p}(q_1)\bs{p}(q'_1))
-\angle(\bs{n}(q_2),\bs{p}(q_2)\bs{p}(q'_2)) \nonumber \\
\Delta_5(c,c') := &\angle (\bs{n}(q'_1),\bs{p}(q_1)\bs{p}(q'_1))-\angle(\bs{n}(q'_2),\bs{p}(q_2)\bs{p}(q'_2)) \nonumber 
\end{align}

\begin{figure}[t]
\centering
\begin{overpic}[width=1.0\columnwidth]{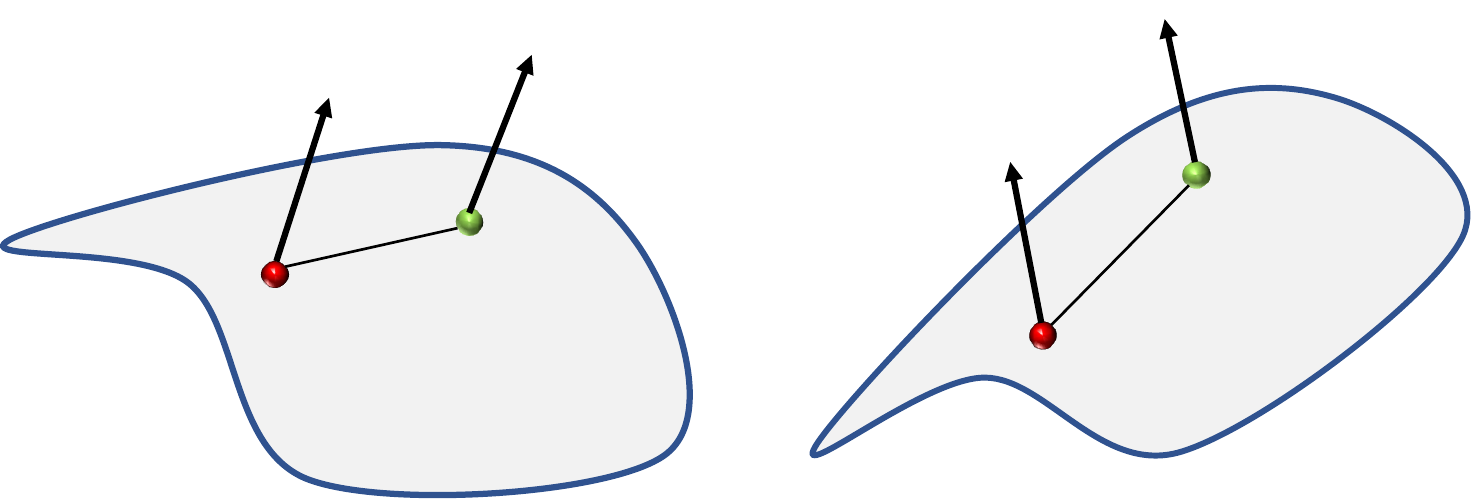}
\put(13,10){$\bs{p}(q_1)$}
\put(35,16){$\bs{p}(q'_1)$}
\put(16,29){$\bs{n}(q_1)$}
\put(33,32){$\bs{n}(q'_1)$}
\put(72,6){$\bs{p}(q_2)$}
\put(85,21){$\bs{p}(q'_2)$}
\put(64,27){$\bs{n}(q_2)$}
\put(80,33){$\bs{n}(q'_2)$}
 \end{overpic}
\caption{The geometry consistency constraints are based on the fact that rigid transform preserves length and angle.}
\label{Figure:Consistency:And:Regression}
\end{figure}
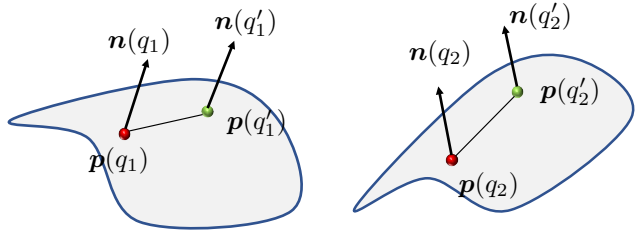

\normalsize
where $\Delta_1(c,c')$ measures the descriptor consistency, and $\Delta_i(c,c'),2\leq i \leq 5$, as motivated by~\cite{Shan:2004:LMH,Huang:2006:RFO}, measure the consistency in edge length and angles (see Figure~\ref{Figure:Consistency:And:Regression}). We now define the weight of $(c,c')$ as
\begin{equation}
w_{\gamma}(c,c') = \exp\Big(-\frac{1}{2}\sum\limits_{i=1}^5\big(\frac{\Delta_i(c,c')}{\gamma_i}\big)^2\Big)
\end{equation}
where $\gamma = (\gamma_1,\gamma_2,\gamma_3,\gamma_4,\gamma_5)$ are hyper-parameters associated with the consistency measures. 

We define the robust rigid regression loss as
$$
\small{
r_{(R,\bs{t})}(c) = \big(\|R\bs{p}(q_1)+\bs{t}-\bs{p}(q_2)\|^2 +  \|R\bs{n}(q_1) - \bs{n}(q_2)\|^2\big)},
$$
\normalsize

We perform alternating maximization to optimize (\ref{Eq:Objective:Function}). When $R$ and $\bs{t}$ are fixed, (\ref{Eq:Objective:Function}) reduces to 
\begin{equation}
\max_{x_{c}} \ \sum\limits_{c,c'} a_{c c'}x_{c} x_{c'} \quad \textup{subject to} \  \sum\limits_{c}x_{c}^2 = 1,
\end{equation}
where $a_{cc'}:= w_{\gamma}(c,c') \big(\delta-r_{(R,\bs{t})}(c)-r_{(R,\bs{t})}(c')\big)$. It is clear that the optimal solution $\{x_c\}$ is given by the maximum eigenvector of $A = (a_{cc'})$. Likewise, when $\{x_{c}\}$ is fixed, (\ref{Eq:Objective:Function}) reduces to
\begin{equation}
\min\limits_{R,\bs{t}}\ \sum\limits_{c\in \set{C}} a_c r_{(R,\bs{t})}(c), \quad a_{c} := x_{c} \sum\limits_{c'\in \set{C}} w_{\gamma}(c,c')x_{c'}.
\label{Eq:R:t:Robust:Regression}
\end{equation}
We solve (\ref{Eq:R:t:Robust:Regression}) using iterative reweighted least squares. The step exactly follows~\cite{Bouaziz:2013:SIC} and is left to Appendix~\ref{Subsec:IRLS}.  
In our implementation, we use $5$ alternating iterations between spectral matching and robust fitting.

Our approach essentially combines the strengths of iterative reweighted least squares (or IRLS) and spectral matching. IRLS is known to be sensitive to large outlier ratios (c.f.\cite{DBLP:conf/ciss/DaubechiesDFG08}). In our formulation, this limitation is addressed by spectral matching, which detects the strongest consistent correspondence subset. On the other hand, spectral matching, which is a relaxation of a binary-integer program, does not offer a clean separation between inliers and outliers. This issue is addressed by using IRLS. 

\subsection{Network Training}
\label{Section:Network:Training}

We train the proposed network by utilizing training data of the form $\set{P}_{\train} = \{(\{(\overline{I}_i, \overline{S}_i^{\star})\},T^{\star})\}$, where each instance collects two input scans, their corresponding completions, and their relative pose. Network training proceeds in two phases. 

\subsubsection{Learning Each Individual Module}

\noindent\textbf{Learning semantic descriptors.} We begin with learning the proposed feature representation. Since color, depth, normals，semantic label are all pre-specified, we only learn the semantic descriptor channels $\bs{f}$. To this end, we first define a contrastive loss on the representation of scan completions for training globally discriminative descriptors: 
\begin{align}
& \set{L}_{des}(\overline{S}_1, \overline{S}_2) :=  \sum\limits_{(q_1,q_2)\in \set{G}(S_1, S_2)}\|\bs{f}(q_1)-\bs{f}(q_2)\|\nonumber \\
& \quad + \sum\limits_{(q_1,q_2)\in \set{N}(S_1, S_2)}\max(0,D -\|\bs{f}(q_1)-\bs{f}(q_2)\|),
\label{Eq:Descriptor:Loss}
\end{align}
where $\set{G}(S_1, S_2)$ and $\set{N}(S_1, S_2)$ collect randomly sampled corresponding point pairs and non-corresponding point pairs between $S_1$ and $S_2$, respectively. $D$ is set to 0.5 in our experiments. We then solve the following optimization problem to learn semantic descriptors:
\begin{equation}
\min\limits_{\theta} \ \sum\limits_{(\{(\overline{I}_i, \overline{S}_i^{\star})\},T^{\star})\in \set{P}_{\train}}\set{L}_{des}(\overline{S}_1, \overline{S}_2)\quad s.t. \quad \bs{f} = \bs{f}_{\theta}    
\end{equation}
where $\bs{f}_{\theta}$ is the feed-forward network introduced in Section~\ref{Section:Feature:Representation}. In our experiments, we train around 100k iterations with batch size 2 using ADAM optimizer~\cite{DBLP:journals/corr/KingmaB14}.

\noindent\textbf{Learning completion modules.} We train the completion network $g_{\phi}$ by combining a regression loss and a contrastive descriptor loss: 

\small{
\begin{align*}
\min\limits_{\phi}& \sum\limits_{(\{(\overline{I}_i, \overline{S}_i^{\star})\},T^{\star})\in \set{P}_{\train}} \underset{T\sim \set{N}(T^{\star},\Sigma)}{E} \Big(\|g_{\phi}(\hat{I}(\overline{I}_1,\overline{I}_2,T))-S_1\|_{\set{F}}^2\nonumber \\
&+ \lambda \set{L}_{des}(\overline{S}_1, g_{\phi}(\hat{I}(\overline{I}_1,\overline{I}_2,T)))\Big)
\end{align*}
}
\normalsize
\noindent ,where $\lambda = 0.01$. $\hat{I}(\overline{I}_1,\overline{I}_2,T)$ denotes the concatenated input of $\overline{I}_1$ and transformed $\overline{I}_2$ using $T$. We train again around 100k iterations with batch size 2 using ADAM optimizer~\cite{DBLP:journals/corr/KingmaB14}.

The motivation of the contrastive descriptor loss is that the completion network does not fit the training data perfectly, and adding this term improves the performance of descriptor matching. Also noted that the input relative pose is not perfect during the execution of the entire network, thus we randomly perturb the relative pose in the neighborhood of each ground-truth for training.

\normalsize

\noindent\textbf{Pre-training relative pose module.} We pre-train the relative pose module using the results of the bi-scan completion module: 
\begin{equation}
\min\limits_{\gamma}\ \sum\limits_{(\{(\overline{I}_i, \overline{S}_i^{\star})\},T^{\star})\in \set{P}_{\train}} \ \|h_{\gamma}(\overline{S}_1, \overline{S}_2)-T^{\star}\|_{\set{F}}^2
\label{Eq:Pretrain:Loss}
\end{equation}
For optimization, we employ finite-difference gradient descent with backtracking line search~\cite{NoceWrig06} for optimization. In our experiments, the training converges in 30 iterations.

\subsubsection{Fine-tuning Relative Pose Module} 

Given the pre-trained individual modules, we could fine-tune the entire network together. However, we find that the training is hard to converge and the test accuracy even drops. Instead, we find that a more effective fine-tuning strategy is to just optimize the relative pose modules. In particular, we allow them to have different hyper-parameters to accommodate specific distributions of the completion results at different layers of the recurrent network. Specifically, let $\gamma$ and $\gamma_t$ be the hyper-parameters of the first pairwise matching module and the pairwise matching module at iteration $t$, respectively. With $T^{t_{\max}}(\overline{I}_1,\overline{I}_2)$ we denote the output of the entire network. We solve the following optimization problem for fine-tuning:
\begin{equation}
\min\limits_{\gamma, \{\gamma_t\}} \sum\limits_{(\{(\overline{I}_i, \overline{S}_i^{\star})\},T^{\star})\in \set{P}_{\train}} \|T^{t_{\max}}(\overline{I}_1,\overline{I}_2)-T^{\star}\|_{\set{F}}^2.  
\label{Eq:Fine:Tune:Loss}
\end{equation}
Similar to (\ref{Eq:Pretrain:Loss}), we again employ finite-difference gradient descent with backtracking line search~\cite{NoceWrig06} for optimization. To stabilize the training, we further employ a layer-wise optimization scheme to solve (\ref{Eq:Fine:Tune:Loss}) sequentially.  In our experiments, the training converges in 20 iterations.
\section{Experimental Results}
\label{Section:Experimental:Results}

In this section, we present an experimental evaluation of the proposed approach. We begin with describing the experimental setup in Section~\ref{Section:Experimental:Setup}. We then present an analysis of the our results in Section~\ref{Section:Analysis:Results}. Finally, we present an ablation study in Section~\ref{Section:Ablation:Study}. Please refer to Appendix~\ref{Section:Additional:Experimental:Results} for more qualitative results and an enriched ablation study. 

\subsection{Experimental Setup}
\label{Section:Experimental:Setup}

\setlength\tabcolsep{1.45pt}
\begin{table*}
\footnotesize

\begin{tabular}{l|ccc|c|ccc|c|ccc|c|ccc|c|ccc|c|ccc|c}
& \multicolumn{8}{c|}{SUNCG} & \multicolumn{8}{c|}{Matterport} & \multicolumn{8}{c}{ScanNet} \\ \hline
&\multicolumn{4}{c|}{Rotation} & \multicolumn{4}{c|}{Trans.} & \multicolumn{4}{c|}{Rotation} & \multicolumn{4}{c|}{Trans.} & \multicolumn{4}{c|}{Rotation} & \multicolumn{4}{c}{Trans.}\\ \hline 
& $\ang{3}$ & $\ang{10}$ & $\ang{45}$ & Mean & $0.1$ & $0.25$& $0.5$ & Mean
& $\ang{3}$ & $\ang{10}$ & $\ang{45}$ & Mean & $0.1$ & $0.25$& $0.5$ & Mean
& $\ang{3}$ & $\ang{10}$ & $\ang{45}$ & Mean & $0.1$ & $0.25$& $0.5$ & Mean
\\ \hline
4PCS([0.5,1]) & 64.3 & 83.7 & 87.6 & 21.0 & 68.2 & 74.4 & 79.0 & 0.30 & 42.7 & 65.7 & 80.3 & 33.4 & 52.6 & 64.3 & 69.0 & 0.46 & 25.3 & 48.7 & 80.1 & 31.2 & 36.9 & 43.2 & 59.8 & 0.52 \\
GReg([0.5,1]) & 85.9 & 91.9 & 94.1 & 10.3 & 86.9 & 89.3 & 90.7 & 0.16 &  80.8 & 89.2 & 92.1 & 12.0 & 84.8 & 88.5 & 90.6 & 0.17 & 58.9 & 84.4 & 88.8 & 16.3 & 81.7 & 85.8 & 88.6 & \textbf{0.19}\\
CGReg([0.5,1]) & 90.8 & 92.9 & 93.9 & 9.8 & 87.3 & 90.7 & 92.8 & 0.13 & 90.3 & 90.8 & 93.1 & 10.1 & 89.4 & 89.6 & 91.6 & 0.14 & 59.0 & 75.7 & 88.1 & 18.0 & 62.1 & 77.7 & 86.9 & 0.23\\

DL([0.5, 1]) & 0.0 & 0.0 & 15.9 & 81.4 & 0.0 & 1.9 & 8.5 & 1.60 & 0.0 & 0.0 & 9.9 & 83.8 & 0.0 & 3.3 & 6.6 & 1.77 & 0.0 & 0.0 & 30.0 & 61.3 & 0.0 & 0.1 & 0.7 & 3.31\\

Ours-nc.([0.5,1]) & 88.6 & 94.7 & 97.6 & 4.3 & 83.4 & 92.6 & 95.9 & \textbf{0.10} & 90.5 & 97.6 & 98.9 & 2.3 & 93.7 & 96.9 & 98.9 & \textbf{0.04} & 57.2 & 80.6 & 90.5 & 13.9 & 66.3 & 79.6 & 85.9 & 0.24 \\
Ours-nr.([0.5,1]) & 90.0 & 96.0 & 97.8 & 4.3 & 83.8 & 94.4 & 96.5 & \textbf{0.10} & 85.9 & 97.7 & 99.0 & 2.7 & 88.9 & 94.6 & 97.2 & 0.07 & 51.0 & 78.3 & 91.2 & \textbf{12.7} & 63.7 & 79.2 & 86.8 & 0.22 \\
Ours([0.5, 1]) & 90.9 & 95.9 & 97.8 & \textbf{4.0} & 83.6 & 94.3 & 96.6 & \textbf{0.10} & 89.5 & 98.5 & 99.3 & \textbf{1.9} & 93.1 & 96.7 & 98.5 & 0.05 & 52.9 & 79.1 & 91.3 & \textbf{12.7} & 64.7 & 78.6 & 86.0 & 0.23 \\

\hline
4PCS([0.1,0.5)) & 4.9 & 10.6 & 13.7 & 113.0 & 4.0 & 5.3 & 7.1 & 1.99 & 4.2 & 16.2 & 25.9 & 87.0 & 5.0 & 8.1 & 10.0 & 2.19 & 1.5 & 7.1 & 30.0 & 82.2 & 2.5 & 3.1 & 3.1 & 1.63\\
GReg([0.1,0.5)) & 35.1 & 45.4 & 50.3 & 64.1 & 35.8 & 40.3 & 43.6 & 1.29 & 19.2 & 26.8 & 34.9 & 73.8 & 24.2 & 27.2 & 28.4 & 1.68 & 11.4 & 25.0 & 33.3 & 86.5 & 18.1 & 21.7 & 23.4 & 1.31\\
CGReg([0.1,0.5]) & 46.4 & 48.5 & 51.0 & 63.4 & 40.2 & 42.7 & 46.0 & 1.34 & 28.5 & 29.3 & 35.9 & 73.9 & 28.1 & 28.3 & 29.5 & 1.99 & 11.8 & 20.0 & 32.9 & 88.2 & 11.6 & 16.0 & 21.0 & 1.36\\

DL([0.1, 0.5)) & 0.0 & 0.0 & 8.0 & 94.0 & 0.0 & 1.8 & 4.0 & 2.06 & 0.0 & 0.0 & 8.5 & 94.3 & 0.0 & 0.4 & 2.7 & 2.25 & 0.0 & 0.0 & 7.5 & 92.1 & 0.0 & 0.0 & 0.0 & 4.03\\

Ours-nc.([0.1,0.5]) & 47.5 & 62.6 & 71.4 & 32.8 & 36.3 & 54.6 & 63.4 & 0.89 & 54.4 & 75.7 & 83.7 & 22.8 & 53.3 & 65.3 & 73.7 & 0.55 & 14.1 & 37.1 & 56.0 & 55.3 & 18.8 & 31.2 & 41.3 & 0.98\\
Ours-nr.([0.1,0.5]) & 60.3 & 80.3 & 83.7 & 20.8 & 41.2 & 70.0 & 80.6 & 0.56 & 47.3 & 72.9 & 82.4 & 24.6 & 44.4 & 65.1 & 73.9 & 0.57 & 12.2 & 36.0 & 65.3 & 45.2 & 18.1 & 33.6 & 47.0 & 0.90\\
Ours([0.1,0.5)) & 67.2 & 84.1 & 86.4 & \textbf{18.1} & 44.8 & 73.8 & 83.9 & \textbf{0.49}
  & 53.7 & 80.7 & 87.9 & \textbf{17.2} & 52.0 & 71.2 & 81.4 & \textbf{0.45}
 & 14.4 & 39.1 & 66.8 & \textbf{43.9} & 19.6 & 35.5 & 48.4 & \textbf{0.87}\\ \hline

DL([0.0, 0.1)) & 0.0 & 0.0 & 2.1 & 115.4 & 0.0 & 1.4 & 4.3 & 2.23 & 0.0 & 0.0 & 2.1 & 125.9 & 0.0 & 0.2 & 2.1 & 2.83 & 0.0 & 0.0 & 0.0 & 130.4 & 0.0 & 0.0 & 0.0 & 5.37\\

Ours-nc.([0.0,0.1]) & 2.2 & 5.8 & 13.8 & 102.1 & 0.1 & 0.7 & 5.6 & 2.21 & 1.3 & 4.9 & 11.7 & 117.3 & 0.2 & 0.2 & 0.9 & 3.10 & 0.5 & 4.8 & 16.3 & 99.4 & 0.0 & 0.5 & 2.2 & 1.92\\
Ours-nr.([0.0,0.1]) & 12.6 & 27.1 & 33.8 & 83.4 & 3.2 & 15.7 & 28.8 & 1.78 & 1.6 & 11.4 & 27.3 & 92.6 & 0.2 & 2.2 & 7.3 & 2.33 & 0.7 & 7.7 & 29.1 & 83.4 & 0.2 & 1.7 & 7.6 & 1.70\\
Ours([0.0,0.1)) & 15.7 & 32.4 & 37.7 & \textbf{79.5} & 4.5 & 21.3 & 34.3 & \textbf{1.66} & 2.5 & 16.3 & 31.3 & \textbf{87.3} & 0.3 & 3.0 & 11.7 & \textbf{2.19}
 & 0.9 & 8.8 & 32.8 & \textbf{78.9} & 0.4 & 2.3 & 8.7 & \textbf{1.62}\\ 
 Identity([0.0,0.1)) &  &  &  & 103.8 &  &  &  & 2.37 &  &  &  & 131.1 &  &  & 
 & 3.20 &  &  &  & 82.5 &  &  & & 1.96\\ 
 \hline

\end{tabular}
\caption{\small{Benchmark evaluation on our approach and baseline approaches. Ours-nc and Ours-nr stand for ours method with completion module and recurrent module removed, respectively. For the rotation component, we show the percentage of pairs whose angular deviations fall within $3^{\circ}$,$10^{\circ}$, and $45^{\circ}$, respectively. For the translation component, we show the percentage of pairs whose translation deviations fall within $0.1m$,$0.25m$,$0.5m$
We also show the mean errors. In addition, we show statistics for pairs of scans whose overlapping ratios fall into three
intervals, namely, [50\%, 100\%], [10\%, 50\%], and [0\%, 10\%]. Average numbers are reported for 10 repeated runs on test sets.}}
\label{Table:Benchmark:Evaluation}
\vspace{-0.05in}
\end{table*}

\subsubsection{Datasets}
We perform experimental evaluation on three datasets:
\noindent\textbf{SUNCG}~\cite{song2016ssc} is a synthetic dataset that collects 45k different 3D scenes, where we take 9892 bedrooms for experiments. For each room, we sample 25 camera locations around the room center, the field of view is set as $90^{\circ}$ horizontally and $90^{\circ}$ vertically. From each camera pose we collect an input scan and the underlying ground-truth completion stored in local coordinate system of that camera pose. We allocate 80\% rooms and the rest for testing. 
\noindent\textbf{Matterport}~\cite{DBLP:journals/corr/abs-1709-06158} is a real dataset that collects 925 different 3D scenes. Each room was reconstructed from a real indoor room. We use their default train/test split. For each room, we pick 50 camera poses. The sampling strategy and camera configuration are the same as SUNCG.
\noindent\textbf{ScanNet}~\cite{DBLP:journals/corr/DaiCSHFN17} is a real dataset that collects 1513 rooms. Each room was reconstructed using thousands of depth scans from Kinect. For each room, we select every 25 frames in the recording sequence. For each camera location, we render the cube-map representation using the reconstructed 3D model. Note that unlike SUNCG and Matterport, where the reconstruction is complete. The reconstruction associated with ScanNet is partial, i.e., there are much more areas in our cube-map representation that are missing values due to the incompleteness of ground truth. For testing, we sample 1000 pair of scans (source and target scan are from the same room) for all datasets.

\subsubsection{Baseline Comparison} 
We consider four baseline approaches:

\noindent\textbf{Super4PCS~\cite{CGF:CGF12446}} is a state-of-the-art non-deep learning technique for relative pose estimation between two 3D point clouds. It relies on using geometric constraints to vote for consistent feature correspondences. We used the author's code for comparison.

\noindent\textbf{Global registration (or GReg)~\cite{DBLP:journals/corr/abs-1801-09847}} is another state-of-the-art non-deep learning technique for relative pose estimation. It combines cutting-edge feature extraction and reweighted least squares for rigid pose registration. GReg is a more robust version than fast global registration (or FGReg)~\cite{DBLP:conf/eccv/ZhouPK16}, which focuses on efficiency. We used the Open3D implementation of GReg for comparison.

\noindent\textbf{Colored Point-cloud Registration (or CGReg)~\cite{Park2017ColoredPC}} This method is a combination of GReg and colored point-cloud registration, where color information is used to boost the accuracy of feature matching. We used the Open3D implementation.

\noindent\textbf{Deep learning baseline (or DL)\cite{melekhov2017relative}} is the most relevant deep learning approach for estimating the relative pose between a pair of scans. It uses a Siamese network to extract features from both scans and regress the quaternion and translation vectors. We use the authors' code and modify their network to take in color, depth, normal as input. Note that we did not directly compare to~\cite{ranftl2018deep} as extending it to compute relative poses between RGB-D scans is non-trivial, and our best attempt was not as competitive as the pairwise matching module introduced in this paper.

\begin{figure*}
\centering
\footnotesize
\def\imh{0.073\textwidth}
\def\imw{0.12\textwidth}
\newcommand{\T}[1]{\raisebox{-0.5\height}{#1}}
\setlength{\tabcolsep}{1pt}
\begin{tabular}{ccccccccc}

\rotatebox[origin=c]{90}{G.T. Color} &
\T{\includegraphics[width=\imw]     {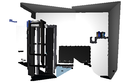}} & 
\T{\includegraphics[width=\imw]   {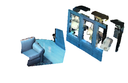}} & 
\T{\includegraphics[width=\imw]      {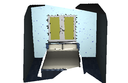}} & 
\T{\includegraphics[width=\imw]    {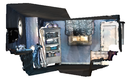}} & 
\T{\includegraphics[width=\imw]      {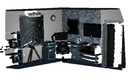}} & 
\T{\includegraphics[width=\imw]     {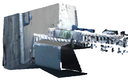}} & 
\T{\includegraphics[width=\imw]    {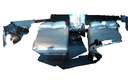}} & 
\T{\includegraphics[width=\imw]       {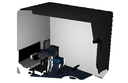}} \\
[+21pt]

\rotatebox[origin=c]{90}{G.T. Scene} &
\T{\includegraphics[width=\imw]     {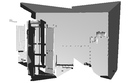}} & 
\T{\includegraphics[width=\imw]   {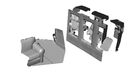}} & 
\T{\includegraphics[width=\imw]      {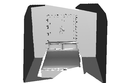}} & 
\T{\includegraphics[width=\imw]    {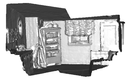}} & 
\T{\includegraphics[width=\imw]      {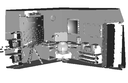}} & 
\T{\includegraphics[width=\imw]     {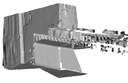}} & 
\T{\includegraphics[width=\imw]    {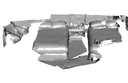}} & 
\T{\includegraphics[width=\imw]       {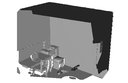}} \\
[+21pt]

\rotatebox[origin=c]{90}{Ours} &
\T{\includegraphics[width=\imw]     {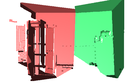}} & 
\T{\includegraphics[width=\imw]   {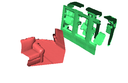}} & 
\T{\includegraphics[width=\imw]      {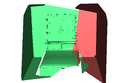}} & 
\T{\includegraphics[width=\imw]    {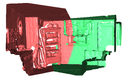}} & 
\T{\includegraphics[width=\imw]      {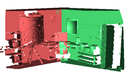}} & 
\T{\includegraphics[width=\imw]     {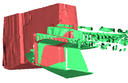}} & 
\T{\includegraphics[width=\imw]    {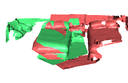}} & 
\T{\includegraphics[width=\imw]       {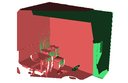}} \\
[+21pt]

\rotatebox[origin=c]{90}{4PCS} &
\T{\includegraphics[width=\imw]       {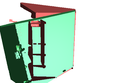}} & 
\T{\includegraphics[width=\imw]     {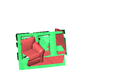}} & 
\T{\includegraphics[width=\imw]     {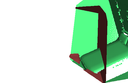}} & 
\T{\includegraphics[width=\imw]{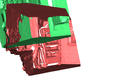}} & 
\T{\includegraphics[width=\imw]     {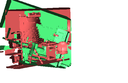}} & 
\T{\includegraphics[width=\imw]    {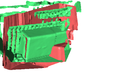}} & 
\T{\includegraphics[width=\imw]   {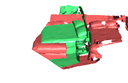}} & 
\T{\includegraphics[width=\imw]      {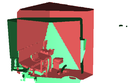}} \\
[+21pt]

\rotatebox[origin=c]{90}{DL} &
\T{\includegraphics[width=\imw]     {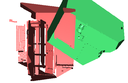}} & 
\T{\includegraphics[width=\imw]   {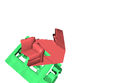}} & 
\T{\includegraphics[width=\imw]      {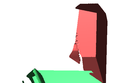}} & 
\T{\includegraphics[width=\imw]    {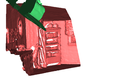}} & 
\T{\includegraphics[width=\imw]      {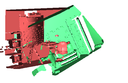}} & 
\T{\includegraphics[width=\imw]     {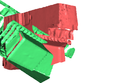}} & 
\T{\includegraphics[width=\imw]    {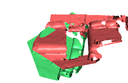}} & 
\T{\includegraphics[width=\imw]       {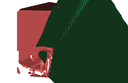}} \\
[+21pt]

\rotatebox[origin=c]{90}{GReg} &
\T{\includegraphics[width=\imw]     {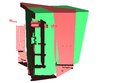}} & 
\T{\includegraphics[width=\imw]   {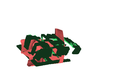}} & 
\T{\includegraphics[width=\imw]      {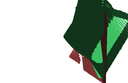}} & 
\T{\includegraphics[width=\imw]    {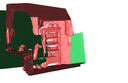}} & 
\T{\includegraphics[width=\imw]      {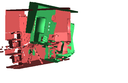}} & 
\T{\includegraphics[width=\imw]     {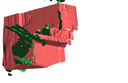}} & 
\T{\includegraphics[width=\imw]    {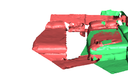}} & 
\T{\includegraphics[width=\imw]       {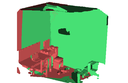}} \\
[+21pt]

\rotatebox[origin=c]{90}{CGReg} &
\T{\includegraphics[width=\imw]     {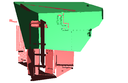}} & 
\T{\includegraphics[width=\imw]   {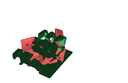}} & 
\T{\includegraphics[width=\imw]      {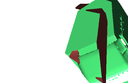}} & 
\T{\includegraphics[width=\imw]    {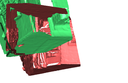}} & 
\T{\includegraphics[width=\imw]      {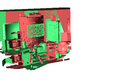}} & 
\T{\includegraphics[width=\imw]     {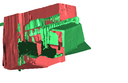}} & 
\T{\includegraphics[width=\imw]    {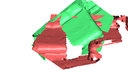}} & 
\T{\includegraphics[width=\imw]       {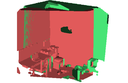}} \\

\rotatebox[origin=c]{90}{G.T. 1} &
\T{\includegraphics[width=\imw]     {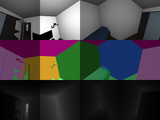}} & 
\T{\includegraphics[width=\imw]   {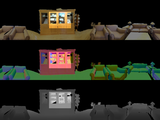}} & 
\T{\includegraphics[width=\imw]      {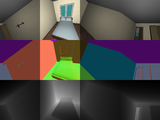}} & 
\T{\includegraphics[width=\imw]    {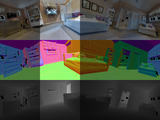}} & 
\T{\includegraphics[width=\imw]      {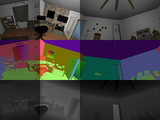}} & 
\T{\includegraphics[width=\imw]     {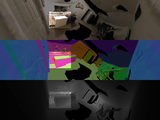}} & 
\T{\includegraphics[width=\imw]    {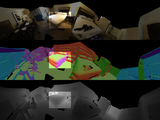}} & 
\T{\includegraphics[width=\imw]       {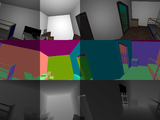}} \\
[+21pt]

\rotatebox[origin=c]{90}{Completed 1} &
\T{\includegraphics[width=\imw]     {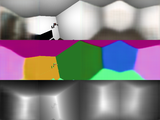}} & 
\T{\includegraphics[width=\imw]   {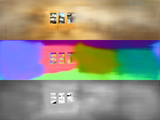}} & 
\T{\includegraphics[width=\imw]      {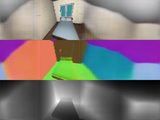}} & 
\T{\includegraphics[width=\imw]    {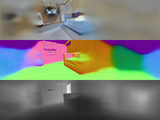}} & 
\T{\includegraphics[width=\imw]      {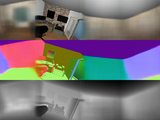}} & 
\T{\includegraphics[width=\imw]     {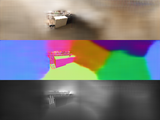}} & 
\T{\includegraphics[width=\imw]    {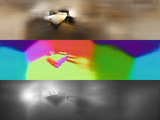}} & 
\T{\includegraphics[width=\imw]       {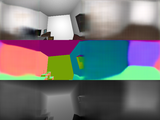}} \\
[+21pt]

\rotatebox[origin=c]{90}{G.T. 2} &
\T{\includegraphics[width=\imw]     {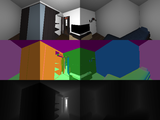}} & 
\T{\includegraphics[width=\imw]   {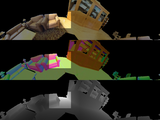}} & 
\T{\includegraphics[width=\imw]      {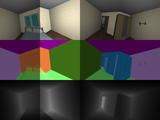}} & 
\T{\includegraphics[width=\imw]    {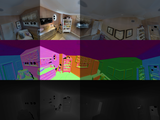}} & 
\T{\includegraphics[width=\imw]      {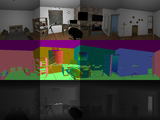}} & 
\T{\includegraphics[width=\imw]     {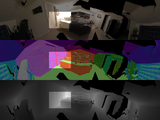}} & 
\T{\includegraphics[width=\imw]    {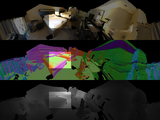}} & 
\T{\includegraphics[width=\imw]       {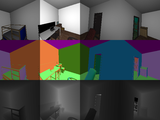}} \\
[+21pt]

\rotatebox[origin=c]{90}{Completed 2} &
\T{\includegraphics[width=\imw]     {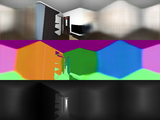}} & 
\T{\includegraphics[width=\imw]   {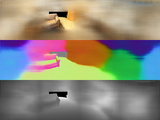}} & 
\T{\includegraphics[width=\imw]      {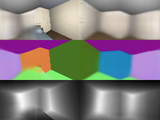}} & 
\T{\includegraphics[width=\imw]    {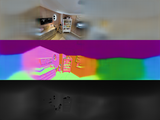}} & 
\T{\includegraphics[width=\imw]      {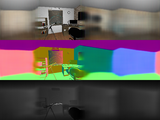}} & 
\T{\includegraphics[width=\imw]     {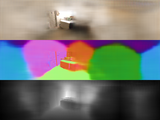}} & 
\T{\includegraphics[width=\imw]    {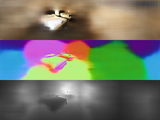}} & 
\T{\includegraphics[width=\imw]       {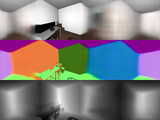}} \\ [-3pt]
& \multicolumn{2}{c}{$\underbrace{\phantom{XXXXXXXXXXXXX}}_\text{No overlap}$} 
& \multicolumn{3}{c}{$\underbrace{\phantom{XXXXXXXXXXXXXXXXX}}_\text{Small overlap}$} 
& \multicolumn{3}{c}{$\underbrace{\phantom{XXXXXXXXXXXXXXXXX}}_\text{Significant overlap}$} \\ [-5pt]
\end{tabular}

\captionof{figure}{\small{Qualitative results of our approach and baseline approaches. We show examples for the cases of no, small and significant overlap. From top to bottom: ground-truth color and scene geometry, our pose estimation results (two input scans in red and green), baseline results (4PCS, DL, GReg and CGReg), ground-truth scene RGBDN and completed scene RGBDN for two input scans. The unobserved regions are dimmed. See Section ~\ref{Section:Analysis:Results} for details. }}
\label{Figure:Visualizations:Results}
\end{figure*}

\subsubsection{Evaluation Protocol} 

We evaluate the rotation component $R$ and translation component $\bs{t}$ of a relative pose $T = (R, \bs{t})$ separately. Let $R^{\star}$ be the ground-truth, we follow the convention of reporting the relative rotation angle $acos(\frac{\|R^{\star}R^{T}\|_{\set{F}}}{\sqrt{2}})$. Let $\bs{t}^{\star}$ be the ground-truth translation. We evaluate the accuracy of $\bs{t}$ by measuring $\|\bs{t}-\bs{t}^{\star}+ (R-R^{\star})\bs{c}_{I_s}\|$, where $\bs{c}_{I_s}$ is the barycenter of $I_s$. 

To understand the behavior of each approach on different types of scan pairs, we divide the scan pairs into three categories. For this purpose, we first define the overlap ratio between a pair of scans $I_s$ and $I_t$ as $o(I_s,I_t) = |I_s\cap I_t|/\min(|I_s|,|I_t|)$. We say a testing pair $(I_s,I_t)$ falls into the category of significant overlap, small overlap, and non-overlap if $o(I_s,I_t)\geq 0.5$, $0.5\geq o(I_s,I_t)\geq 0.1$, and $o(I_s,I_t) \leq 0.1$, respectively.

\subsection{Analysis of Results}
\label{Section:Analysis:Results}

Table~\ref{Table:Benchmark:Evaluation} and Figure~\ref{Figure:Visualizations:Results} provide quantitative and qualitative results of our approach and baseline approaches. Overall, our approach outputs accurate relative pose estimations. The predicted normal are more accurate than color and depth. 

In the following, we provide a detailed analysis under each category of scan pairs as well as the scan completion results:

\noindent\textbf{Significant overlap.} Our approach outputs accurate relative poses in the presence of significant overlap. The mean error in rotation/translation of our approach is $3.9^{\circ}/0.10m$, $1.8^{\circ}/0.05m$, and $13.0^{\circ}/0.23m$ on SUNCG, Matterport, and ScanNet, respectively, In contrast, the mean error in rotation/translation of the top performing methods only achieve $9.8^{\circ}/0.13m$, $10.1^{\circ}/0.14m$, and $16.3^{\circ}/0.19m$, respectively. Meanwhile, the performance of our method drops when the completion component is removed. This means that although there are rich features to match between significantly overlapping scans, performing scan completion still matters. Moreover, our approach achieves better relative performance on SUNCG and Matterport, as the field-of-view is wider than ScanNet. 

\noindent\textbf{Small overlap.} Our approach outputs good relative poses in the presence of small overlap. The mean errors in rotation/translation of our approach are $20.1^{\circ}/0.52m$, $16.3^{\circ}/0.45m$, and $47.4^{\circ}/0.90m$ on SUNCG, Matterport, and ScanNet, respectively. In contrast, the top-performing method only achieves mean errors $63.4^{\circ}/1.29m$, $73.8^{\circ}/1.68m$, and $82.2^{\circ}/1.31m$, leaving a big margin from our approach. Moreover, the relative improvements 

are more salient than scan pairs that possess significant overlaps.

This is expected as there are less features to match from the original scans, and scan completion provides more features to match.  

\begin{figure}[t]
\centering
\includegraphics[width=0.155\textwidth]{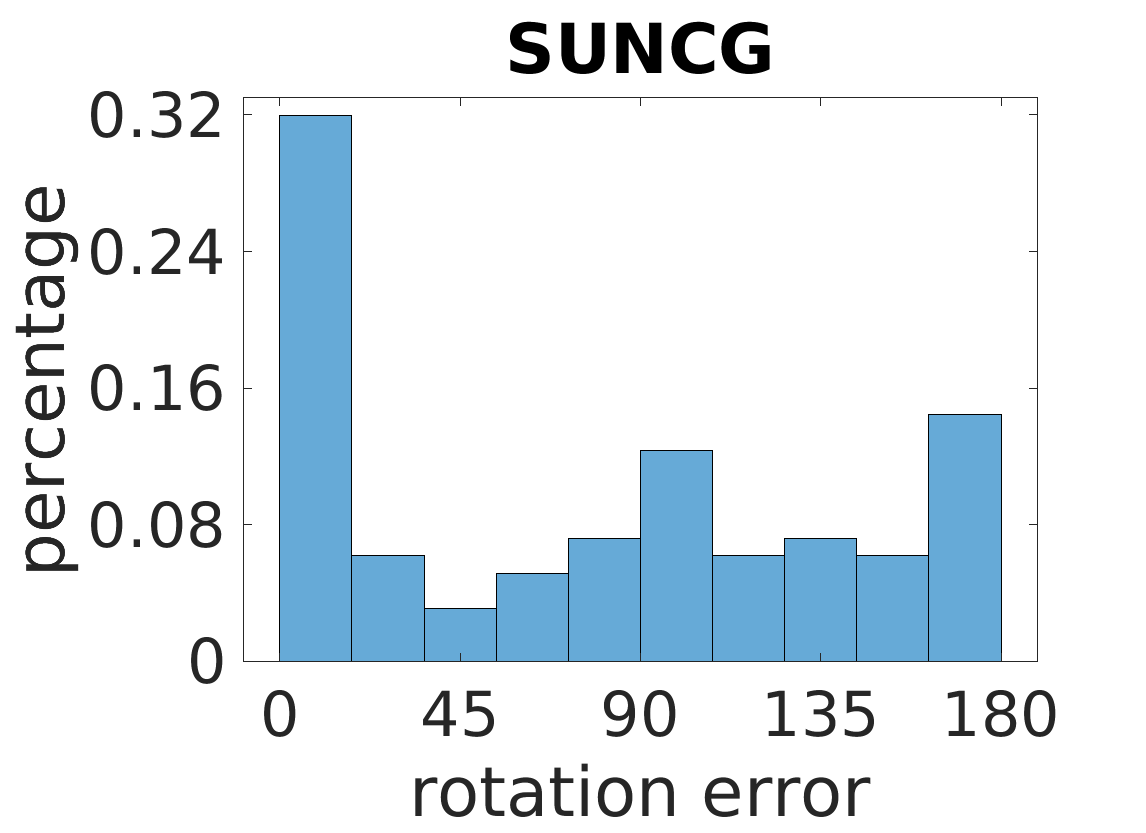}
\includegraphics[width=0.155\textwidth]{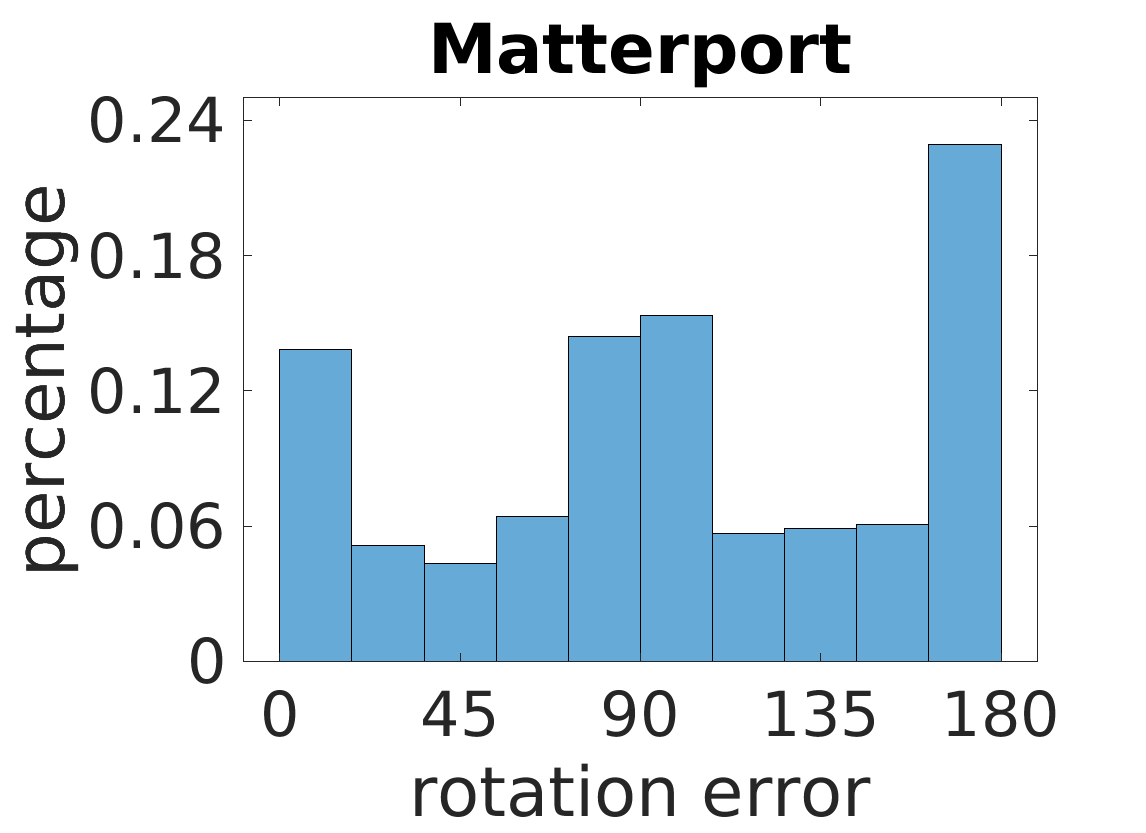}
\includegraphics[width=0.155\textwidth]{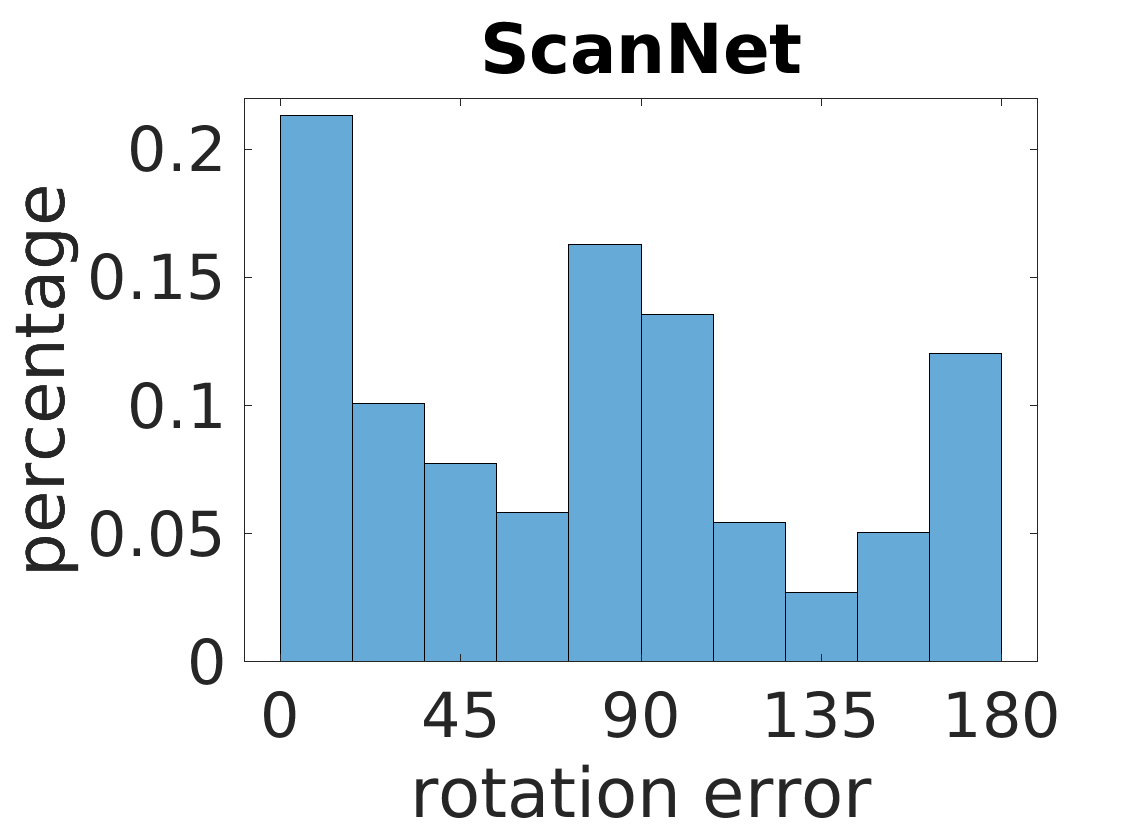}
\caption{Error distribution of rotation errors of our approach on non-overlapping scans. See Section ~\ref{Section:Analysis:Results} for discussion. }
\label{Figure:Error:Distribution}
\end{figure}
\begin{figure}[t]
\centering
\includegraphics[width=0.23\textwidth]{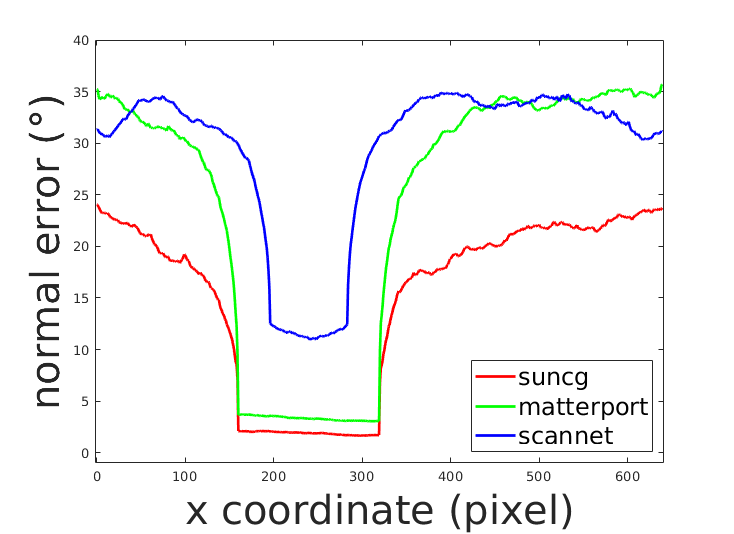}
\includegraphics[width=0.23\textwidth]{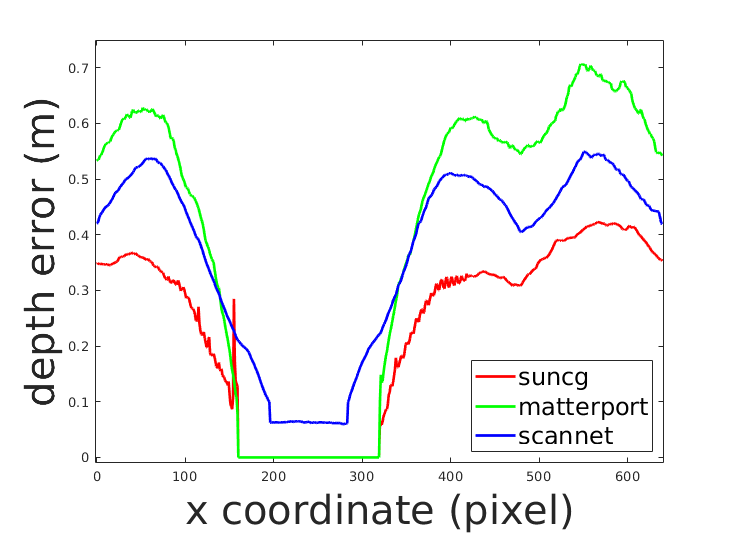}
\caption{Mean errors in predicted normal and depth w.r.t the horizontal image coordinate. See Section ~\ref{Section:Analysis:Results} for discussion.}
\label{Figure:Scan:Completion:Error:Distribution}
\end{figure}

\noindent\textbf{No overlap.} Our approach delivers encouraging relative pose estimations on the extreme non-overlapping scans. For example, in the first column of Figure~\ref{Figure:Visualizations:Results}, a television is separated into two part in source and target scans. Our method correctly assembles the two scans to form a complete scene. In the second example, our method correctly predict the relative position of sofa and a bookshelf.
 
We also show the result (Identity) if we predict identity matrix for each scan pair, which is usually the best we can do for non-overlap scans using traditional method. To further understand our approach, Figure~\ref{Figure:Error:Distribution} plots the error distribution of rotations on the three datasets. We can see a significant portion of the errors concentrate on $90^{\circ}$ and $180^{\circ}$, which can be understood from the perspective that our approach mixes different walls when performing pairwise matching. This is an expected behavior as many indoor rooms are symmetric. Note that we neglect the quantitative results for Super4PCS, GReg, and CGRreg since they all require overlap. 

\noindent\textbf{Scan-completion results.} Figure~\ref{Figure:Scan:Completion:Error:Distribution} plots the error distributions of predicted depth, normal with respect to the horizontal image coordinate. None that in our experiment the $[160,320]$ region is observed for SUNCG/Matterport, and $[196,284]$ for ScanNet. We can see that the errors are highly correlated with the distances to observed region, i.e., they are small in adjacent regions, and become less accurate when the distances become large. This explains why our approach leads to a significant boost on scan pairs with small overlaps, i.e., corresponding points are within adjacent regions.

\subsection{Ablation Study}
\label{Section:Ablation:Study}

We consider two experiments to evaluate the effectiveness of the proposed network design. Each experiment removes one functional unit in the proposed network design. 
\noindent\textbf{No completion.} The first ablation experiment simply applies our relative pose estimation module on the input scans directly, i.e., without scan completions. The performance of our approach drops even on largely overlapping scans, 

This means that it is important to perform scan completions even for partially overlapping scans. Moreover, without completion, our relative pose estimation module still possesses noticeable performance gains against the top-performing baseline GReg~\cite{DBLP:journals/corr/abs-1801-09847} on overlapping scans. Such improvements mainly come from combing spectral matching and robust fitting. Please refer to Appendix~\ref{Section:Additional:Experimental:Results} for in-depth comparison.

\noindent\textbf{No recurrent module.} The second ablation experiment removes the recurrent module in our network design. This reduced network essentially performs scan completion from each input scan and then estimates the relative poses between the scan completions. We can see that the performance drops in almost all the configurations.

This shows the importance of the recurrent module, which leverages bi-scan completions to gradually improve the relative pose estimations.  
\section{Conclusions}
\label{Section:Conclusions}
We introduced an approach for relative pose estimation between a pair of RGB-D scans of the same indoor environment. The key idea of our approach is to perform scan completion to obtain the underlying geometry, from which we then compute the relative pose. Experimental results demonstrated the usefulness of our approach both in terms of its absolute performance when compared to existing approaches and the effectiveness of each module of our approach. In particular, our approach delivers encouraging relative pose estimations between extreme non-overlapping scans. 

{\small
\bibliographystyle{ieee}

}

\appendix
\section{More Technical Details about Our Approach}
\label{Section:More:Technical:Details}

\subsection{Completion Network Architecture}
The completion network takes two sets of RGB-D-N (RGB, depth, and normal) as input. Three separate layers of convolution (followed by ReLU and Batchnorm) are applied to extract domain specific signal before merging. Those three preprocessing-branches are applied to both sets of RGB-D-N input. We also use skip layer to facilitate training. The overall architecture is listed as follows, where C(m,n) specify the convolution layer input/output channel. \\

\begin{figure}
\centering
\includegraphics[width=0.5\columnwidth]{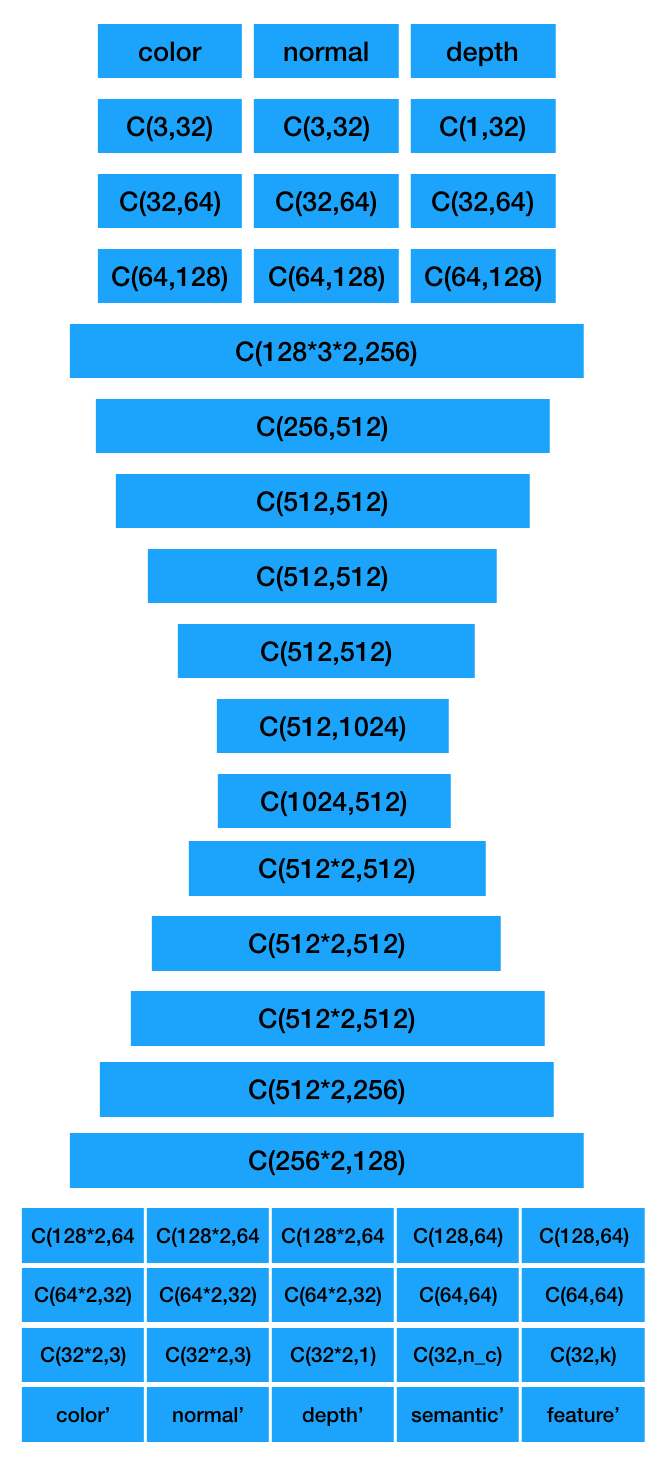}
\caption{Completion network architecture. C(m,n) stands for m input channel and m output channel. Skip connections are added at mirroring location of the encoder and decoder network. Two sets of input(corresponding to source and transformed target scans respectively) first go through the first three layers separately, then being concatenated and pass through the rest layers.}
\label{Figure:Visualizations:supp:network}
\vspace{-0.1in}
\end{figure}

\setlength\tabcolsep{1.45pt}
\begin{table*}[t]
\centering
\footnotesize
\begin{tabular}{l|c|c|c|c|c|c|c|c|c|c|c|c}
& \multicolumn{4}{c|}{SUNCG} & \multicolumn{4}{c|}{Matterport} & \multicolumn{4}{c}{ScanNet} \\ \hline
&\multicolumn{2}{c|}{Rotation} & \multicolumn{2}{c|}{Trans.} & \multicolumn{2}{c|}{Rotation} & \multicolumn{2}{c|}{Trans.} & \multicolumn{2}{c|}{Rotation} & \multicolumn{2}{c}{Trans.}\\ \hline 
& Median & Mean & Median & Mean
& Median & Mean & Median & Mean
& Median & Mean & Median & Mean
\\ \hline
nr & 4.51 & 26.25 & 0.21 & 0.62 & 4.85 & 22.33 & 0.22 & 0.60 & 12.90 & 33.89 & 0.36 & 0.61  \\\hline
r & 1.54 & 23.36 & 0.10 & 0.54 & 2.51 & 18.69 & 0.10 & 0.49 & 7.11 & \textbf{30.40} & 0.23 & \textbf{0.57}  \\\hline
sm & 2.65 & 25.6 & 0.18 & 0.64 & 3.15 & 20.23 & 0.20 & 0.60 & 7.10 & 35.32 & 0.17 & \textbf{0.57}  \\\hline
r+sm & \textbf{1.32} & \textbf{19.36} & \textbf{0.06} & \textbf{0.48} & \textbf{1.45} & \textbf{13.9} & \textbf{0.04} & \textbf{0.34} & \textbf{5.47} & 32.38 & \textbf{0.12} & \textbf{0.57}  \\\hline

\end{tabular}
\caption{\small{Ablation study for pairwise matching. 
nr: Directly apply the closed-form solution~\cite{Horn87closed-formsolution} without reweighted procedure. r: reweighted least square, sm: spectral method, r+sm: alternate between reweighted least square and spectral method. }}
\label{Table:Benchmark:PairWiseMatching}
\vspace{-0.05in}
\end{table*}

{

\subsection{Iteratively Reweighted Least Squares for Solving the Robust Regression Problem}
\label{Subsec:IRLS}

In this section, we provide technical details on solving the following robust regression problem:
\begin{align}
R^{\star}, \bs{t}^{\star} = \underset{R, \bs{t}}{\textup{argmin}}& \sum\limits_{c = (q_1,q_2)} a_{c} \big(\|R \bs{p}(q_1)+\bs{t}-\bs{p}(q_2)\|^2 \nonumber \\
&+ \|R \bs{n}(q_1)-\bs{n}(q_2)\|^2\big)^{\alpha}
\label{Eq:R:t:Least:Square:Regression}
\end{align}
,where we use $\alpha = 1$ in all of our experiments. We solve (\ref{Eq:R:t:Least:Square:Regression}) using reweighted non-linear least squares. Introduce an initial weight $w_c^{(0)} = a_c, c\in \set{C}$. At each iteration $k\geq 0$, we first solve the following non-linear least squares:
\begin{align}
\min\limits_{R,\bs{t}}\ \sum\limits_{c = (q_1,q_2)\in \set{C}} w_c^{(k)}& \big(\|R \bs{p}(q_1)+\bs{t}-\bs{p}(q_2) \|^2 \nonumber \\
&+ \|R \bs{n}(q_1)-\bs{n}(q_2)\|^2\big).
\label{Eq:R:t:Least:Square:Regression2}
\end{align}
According to~\cite{Horn87closed-formsolution}, (\ref{Eq:R:t:Least:Square:Regression}) admits a closed-form solution. Specifically, define
\begin{align*}
\bs{c}^{(k)}(Q_1)& :=\frac{\sum\limits_{c = (q_1,q_2)\in \set{C}} w_c^{(k)}\bs{p}(q_1)}{\sum\limits_{c = (q_1,q_2)\in \set{C}} w_c^{(k)}}, \quad \\
\bs{c}^{(k)}(Q_2)& :=\frac{\sum\limits_{c = (q_1,q_2)\in \set{C}} w_c^{(k)}\bs{p}(q_2)}{\sum\limits_{c = (q_1,q_2)\in \set{C}} w_c^{(k)}}.
\end{align*}
The optimal translation and rotation to (\ref{Eq:R:t:Least:Square:Regression2}) are given by
$$
\bs{t}^{\star} = \bs{c}^{(k)}(Q_2) - R^{\star}\cdot \bs{c}^{(k)}(Q_1), \ R^{\star} = U\diag(1,1,\textup{sign}(M))V^T,
$$
where $U$ and $V$ are given by the singular value decompostion of
$$
M = U\Sigma V^{T} =  \sum\limits_{(q_1,q_2)\in \set{C}}w_c^{(k)}\big(\overline{\bs{p}}(q_1)\overline{\bs{p}}(q_1)^{T} + \overline{\bs{n}}(q_1)\overline{\bs{n}}(q_1)^{T}\big),
$$
and where
$$
\overline{\bs{p}}(q_1) = \bs{p}(q_1) - \bs{c}^{(k)}(Q_1), \ \overline{\bs{p}}(q_2) = \bs{p}(q_2) - \bs{c}^{(k)}(Q_2).
$$
After obtaining the new optimal transformation $R^{\star}, \bs{t}^{\star}$, we update the weight $w_c^{(k+1)}$ associated with correspondence $c$ at iteration $k+1$ as $w_c^{(k+1)}:= $
$$
\frac{1}{(\epsilon^2 + \|R \bs{p}(q_1)+\bs{t}-\bs{p}(q_2) \|^2 + \|R \bs{n}(q_1)-\bs{n}(q_2)\|^2)^{2-\alpha}}
$$
where $\epsilon$ is a small constant to address the issue of division by zero. 

In our experiments, we used 5 reweighting operations for solving (\ref{Eq:R:t:Least:Square:Regression}).

\subsection{Implementation Details}

\noindent\textbf{Implementation details of the completion network.} We used a combination of 5 source of information(color,normal,depth,semantic label,feature) to supervise the completion network. Specifically, we use $$loss_{recon} = \lambda_c loss_{c}+\lambda_n loss_{n}+\lambda_d loss_{d}+\lambda_s loss_{s}+\lambda_f loss_{f}$$, where we use $l_1$ loss for color, normal, depth, $l_2$ loss for feature, and cross-entropy loss for semantic label. We use $\lambda_c,\lambda_n,\lambda_d,\lambda_f = 1$, $\lambda_s = 0.1$. We trained for 100k iterations using a single GTX 1080Ti. We use Adam optimizer with initial learning rate 0.0002.

\section{Additional Experimental Results}
\label{Section:Additional:Experimental:Results}

Figure~\ref{Figure:Visualizations:Results:SUNCG},~\ref{Figure:Visualizations:Results:Matterport},~\ref{Figure:Visualizations:Results:ScanNet} show more qualitative results on SUNCG, Matterport, and ScanNet, respectively. Table~\ref{Table:Benchmark:PairWiseMatching} gives a detailed ablation study of our proposed pairwise matching algorithm. We compare against three variants, namely, direct regression(nr) using \cite{Horn87closed-formsolution}, reweighted least squares(r) (using the robust norm), and merely using spectral matching (sm). We can see that the combination of reweighted least squares and spectral matching gives the best result. 

We also applied the idea of learning weights for correspondence from data~\cite{ranftl2018deep}. Since~\cite{ranftl2018deep} addresses a different problem of estimating the functional matrix between a pair of RGB images, we tried applying the idea on top of reweighted least squares (r) of our approach, namely, by replacing the reweighting scheme described in Section~\ref{Subsec:IRLS} by a small network for predicting the correspondence weight. However, we find this approach generalized poorly on testing data. In contrast, we found that the spectral matching approach, which leverages geometric constraints that are specifically designed for matching 3D data, leads to additional boost in performance.

\begin{figure*}
\centering
\footnotesize
\def\imh{0.073\textwidth}
\def\imw{0.12\textwidth}
\newcommand{\T}[1]{\raisebox{-0.5\height}{#1}}
\setlength{\tabcolsep}{1pt}
\begin{tabular}{ccccccccc}

\rotatebox[origin=c]{90}{G.T. Color} &
\T{\includegraphics[width=\imw]     {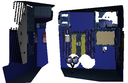}} & 
\T{\includegraphics[width=\imw]   {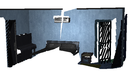}} & 
\T{\includegraphics[width=\imw]      {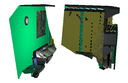}} & 
\T{\includegraphics[width=\imw]    {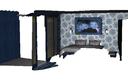}} & 
\T{\includegraphics[width=\imw]      {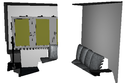}} & 
\T{\includegraphics[width=\imw]     {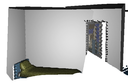}} & 
\T{\includegraphics[width=\imw]    {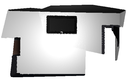}} & 
\T{\includegraphics[width=\imw]       {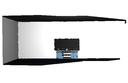}} \\
[+21pt]

\rotatebox[origin=c]{90}{G.T. Scene} &
\T{\includegraphics[width=\imw]     {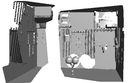}} & 
\T{\includegraphics[width=\imw]   {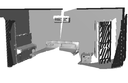}} & 
\T{\includegraphics[width=\imw]      {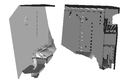}} & 
\T{\includegraphics[width=\imw]    {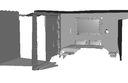}} & 
\T{\includegraphics[width=\imw]      {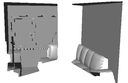}} & 
\T{\includegraphics[width=\imw]     {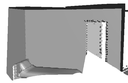}} & 
\T{\includegraphics[width=\imw]    {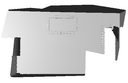}} & 
\T{\includegraphics[width=\imw]       {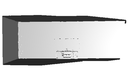}} \\
[+21pt]

\rotatebox[origin=c]{90}{Ours} &
\T{\includegraphics[width=\imw]     {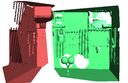}} & 
\T{\includegraphics[width=\imw]   {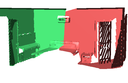}} & 
\T{\includegraphics[width=\imw]      {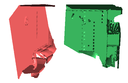}} & 
\T{\includegraphics[width=\imw]    {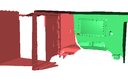}} & 
\T{\includegraphics[width=\imw]      {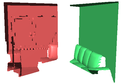}} & 
\T{\includegraphics[width=\imw]     {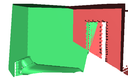}} & 
\T{\includegraphics[width=\imw]    {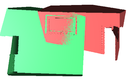}} & 
\T{\includegraphics[width=\imw]       {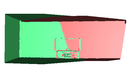}} \\
[+21pt]

\rotatebox[origin=c]{90}{4PCS} &
\T{\includegraphics[width=\imw]     {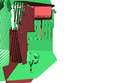}} & 
\T{\includegraphics[width=\imw]   {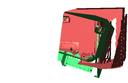}} & 
\T{\includegraphics[width=\imw]      {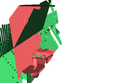}} & 
\T{\includegraphics[width=\imw]    {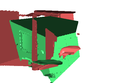}} & 
\T{\includegraphics[width=\imw]      {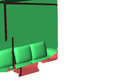}} & 
\T{\includegraphics[width=\imw]     {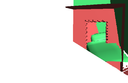}} & 
\T{\includegraphics[width=\imw]    {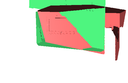}} & 
\T{\includegraphics[width=\imw]       {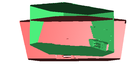}} \\
[+21pt]

\rotatebox[origin=c]{90}{DL} &
\T{\includegraphics[width=\imw]     {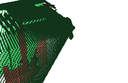}} & 
\T{\includegraphics[width=\imw]   {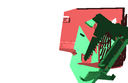}} & 
\T{\includegraphics[width=\imw]      {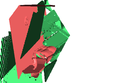}} & 
\T{\includegraphics[width=\imw]    {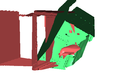}} & 
\T{\includegraphics[width=\imw]      {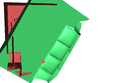}} & 
\T{\includegraphics[width=\imw]     {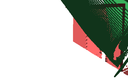}} & 
\T{\includegraphics[width=\imw]    {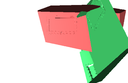}} & 
\T{\includegraphics[width=\imw]       {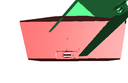}} \\
[+21pt]

\rotatebox[origin=c]{90}{GReg} &
\T{\includegraphics[width=\imw]     {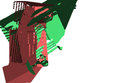}} & 
\T{\includegraphics[width=\imw]   {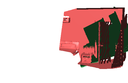}} & 
\T{\includegraphics[width=\imw]      {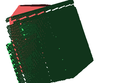}} & 
\T{\includegraphics[width=\imw]    {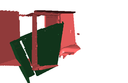}} & 
\T{\includegraphics[width=\imw]      {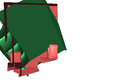}} & 
\T{\includegraphics[width=\imw]     {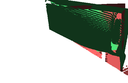}} & 
\T{\includegraphics[width=\imw]    {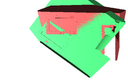}} & 
\T{\includegraphics[width=\imw]       {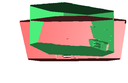}} \\
[+21pt]

\rotatebox[origin=c]{90}{CGReg} &
\T{\includegraphics[width=\imw]     {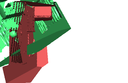}} & 
\T{\includegraphics[width=\imw]   {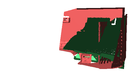}} & 
\T{\includegraphics[width=\imw]      {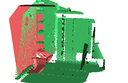}} & 
\T{\includegraphics[width=\imw]    {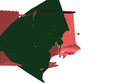}} & 
\T{\includegraphics[width=\imw]      {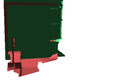}} & 
\T{\includegraphics[width=\imw]     {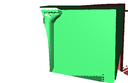}} & 
\T{\includegraphics[width=\imw]    {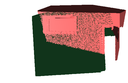}} & 
\T{\includegraphics[width=\imw]       {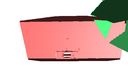}} \\
[+21pt]

\rotatebox[origin=c]{90}{G.T. 1} &
\T{\includegraphics[width=\imw]     {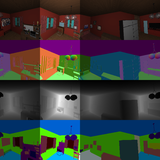}} & 
\T{\includegraphics[width=\imw]   {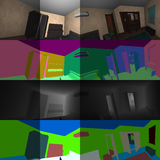}} & 
\T{\includegraphics[width=\imw]      {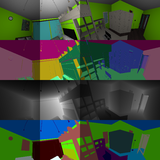}} & 
\T{\includegraphics[width=\imw]    {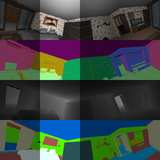}} & 
\T{\includegraphics[width=\imw]      {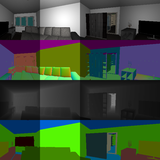}} & 
\T{\includegraphics[width=\imw]     {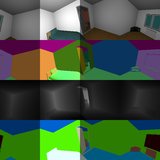}} & 
\T{\includegraphics[width=\imw]    {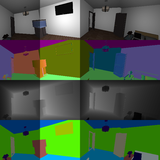}} & 
\T{\includegraphics[width=\imw]       {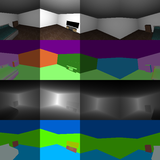}} \\
[+21pt]

\rotatebox[origin=c]{90}{Completed 1} &
\T{\includegraphics[width=\imw]     {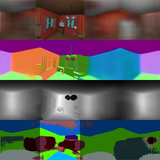}} & 
\T{\includegraphics[width=\imw]   {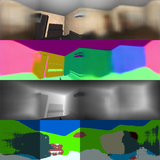}} & 
\T{\includegraphics[width=\imw]      {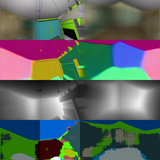}} & 
\T{\includegraphics[width=\imw]    {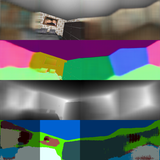}} & 
\T{\includegraphics[width=\imw]      {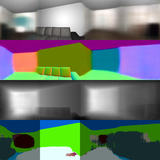}} & 
\T{\includegraphics[width=\imw]     {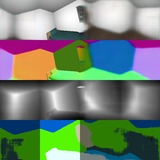}} & 
\T{\includegraphics[width=\imw]    {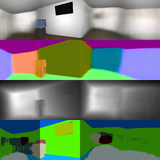}} & 
\T{\includegraphics[width=\imw]       {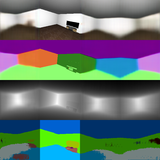}} \\ 
[+21pt]

\rotatebox[origin=c]{90}{G.T. 2} &
\T{\includegraphics[width=\imw]     {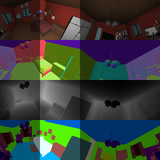}} & 
\T{\includegraphics[width=\imw]   {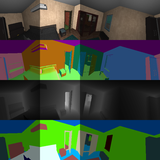}} & 
\T{\includegraphics[width=\imw]      {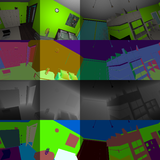}} & 
\T{\includegraphics[width=\imw]    {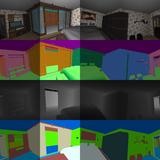}} & 
\T{\includegraphics[width=\imw]      {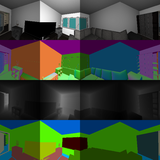}} & 
\T{\includegraphics[width=\imw]     {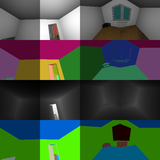}} & 
\T{\includegraphics[width=\imw]    {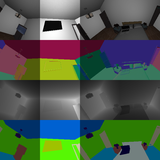}} & 
\T{\includegraphics[width=\imw]       {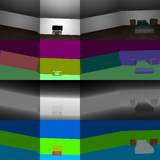}} \\
[+21pt]

\rotatebox[origin=c]{90}{Completed 2} &
\T{\includegraphics[width=\imw]     {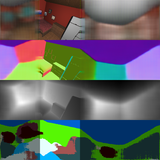}} & 
\T{\includegraphics[width=\imw]   {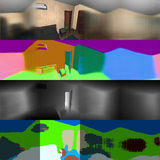}} & 
\T{\includegraphics[width=\imw]      {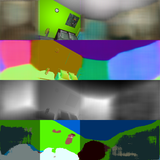}} & 
\T{\includegraphics[width=\imw]    {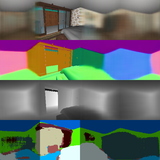}} & 
\T{\includegraphics[width=\imw]      {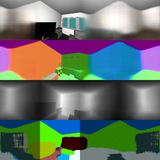}} & 
\T{\includegraphics[width=\imw]     {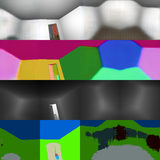}} & 
\T{\includegraphics[width=\imw]    
{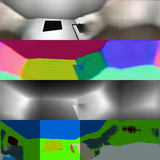}} & 
\T{\includegraphics[width=\imw]       {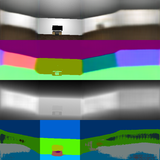}}\\ [-3pt]
\end{tabular}

\captionof{figure}{\small{SUNCG qualitative results. From top to bottom: ground-truth color and scene geometry, our pose estimation results (two input scans in red and green), baseline results (4PCS, DL, GReg and CGReg), ground-truth scene RGBDNS and completed scene RGBDNS for two input scans. The unobserved regions are dimmed.}}
\label{Figure:Visualizations:Results:SUNCG}
\end{figure*}

\begin{figure*}
\centering
\footnotesize
\def\imh{0.073\textwidth}
\def\imw{0.12\textwidth}
\newcommand{\T}[1]{\raisebox{-0.5\height}{#1}}
\setlength{\tabcolsep}{1pt}
\begin{tabular}{ccccccccc}

\rotatebox[origin=c]{90}{G.T. Color} &
\T{\includegraphics[width=\imw]     {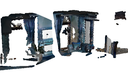}} & 
\T{\includegraphics[width=\imw]   {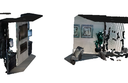}} & 
\T{\includegraphics[width=\imw]      {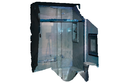}} & 
\T{\includegraphics[width=\imw]    {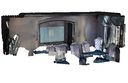}} & 
\T{\includegraphics[width=\imw]      {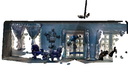}} & 
\T{\includegraphics[width=\imw]     {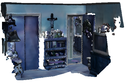}} & 
\T{\includegraphics[width=\imw]    {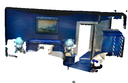}} & 
\T{\includegraphics[width=\imw]       {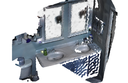}} \\
[+21pt]

\rotatebox[origin=c]{90}{G.T. Scene} &
\T{\includegraphics[width=\imw]     {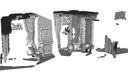}} & 
\T{\includegraphics[width=\imw]   {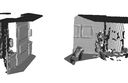}} & 
\T{\includegraphics[width=\imw]      {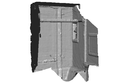}} & 
\T{\includegraphics[width=\imw]    {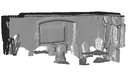}} & 
\T{\includegraphics[width=\imw]      {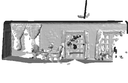}} & 
\T{\includegraphics[width=\imw]     {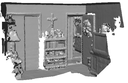}} & 
\T{\includegraphics[width=\imw]    {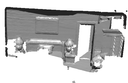}} & 
\T{\includegraphics[width=\imw]       {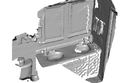}} \\
[+21pt]

\rotatebox[origin=c]{90}{Ours} &
\T{\includegraphics[width=\imw]     {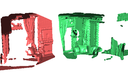}} & 
\T{\includegraphics[width=\imw]   {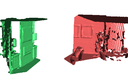}} & 
\T{\includegraphics[width=\imw]      {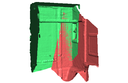}} & 
\T{\includegraphics[width=\imw]    {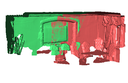}} & 
\T{\includegraphics[width=\imw]      {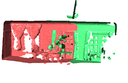}} & 
\T{\includegraphics[width=\imw]     {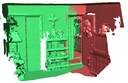}} & 
\T{\includegraphics[width=\imw]    {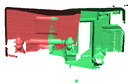}} & 
\T{\includegraphics[width=\imw]       {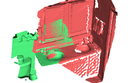}} \\
[+21pt]

\rotatebox[origin=c]{90}{4PCS} &
\T{\includegraphics[width=\imw]     {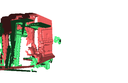}} & 
\T{\includegraphics[width=\imw]   {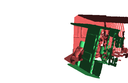}} & 
\T{\includegraphics[width=\imw]      {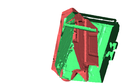}} & 
\T{\includegraphics[width=\imw]    {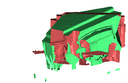}} & 
\T{\includegraphics[width=\imw]      {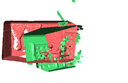}} & 
\T{\includegraphics[width=\imw]     {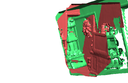}} & 
\T{\includegraphics[width=\imw]    {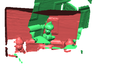}} & 
\T{\includegraphics[width=\imw]       {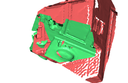}} \\
[+21pt]

\rotatebox[origin=c]{90}{DL} &
\T{\includegraphics[width=\imw]     {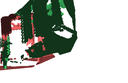}} & 
\T{\includegraphics[width=\imw]   {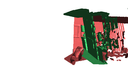}} & 
\T{\includegraphics[width=\imw]      {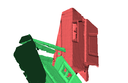}} & 
\T{\includegraphics[width=\imw]    {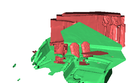}} & 
\T{\includegraphics[width=\imw]      {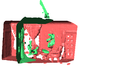}} & 
\T{\includegraphics[width=\imw]     {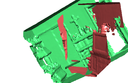}} & 
\T{\includegraphics[width=\imw]    {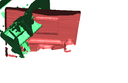}} & 
\T{\includegraphics[width=\imw]       {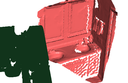}} \\
[+21pt]

\rotatebox[origin=c]{90}{GReg} &
\T{\includegraphics[width=\imw]     {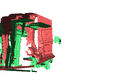}} & 
\T{\includegraphics[width=\imw]   {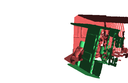}} & 
\T{\includegraphics[width=\imw]      {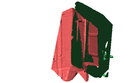}} & 
\T{\includegraphics[width=\imw]    {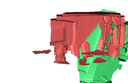}} & 
\T{\includegraphics[width=\imw]      {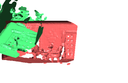}} & 
\T{\includegraphics[width=\imw]     {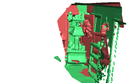}} & 
\T{\includegraphics[width=\imw]    {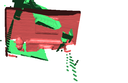}} & 
\T{\includegraphics[width=\imw]       {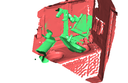}} \\
[+21pt]

\rotatebox[origin=c]{90}{CGReg} &
\T{\includegraphics[width=\imw]     {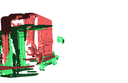}} & 
\T{\includegraphics[width=\imw]   {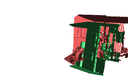}} & 
\T{\includegraphics[width=\imw]      {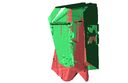}} & 
\T{\includegraphics[width=\imw]    {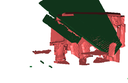}} & 
\T{\includegraphics[width=\imw]      {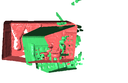}} & 
\T{\includegraphics[width=\imw]     {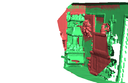}} & 
\T{\includegraphics[width=\imw]    {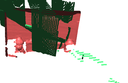}} & 
\T{\includegraphics[width=\imw]       {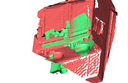}} \\
[+21pt]

\rotatebox[origin=c]{90}{G.T. 1} &
\T{\includegraphics[width=\imw]     {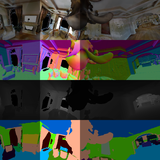}} & 
\T{\includegraphics[width=\imw]   {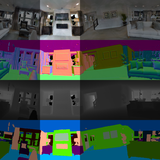}} & 
\T{\includegraphics[width=\imw]      {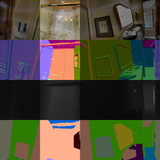}} & 
\T{\includegraphics[width=\imw]    {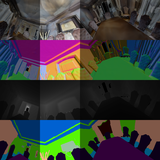}} & 
\T{\includegraphics[width=\imw]      {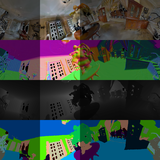}} & 
\T{\includegraphics[width=\imw]     {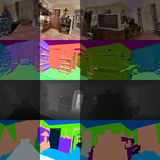}} & 
\T{\includegraphics[width=\imw]    {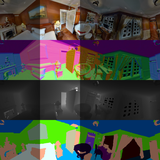}} & 
\T{\includegraphics[width=\imw]       {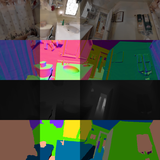}} \\
[+21pt]

\rotatebox[origin=c]{90}{Completed 1} &
\T{\includegraphics[width=\imw]     {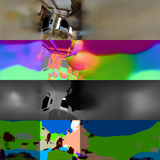}} & 
\T{\includegraphics[width=\imw]   {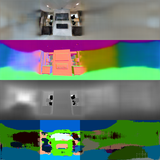}} & 
\T{\includegraphics[width=\imw]      {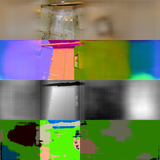}} & 
\T{\includegraphics[width=\imw]    {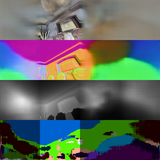}} & 
\T{\includegraphics[width=\imw]      {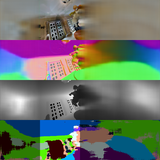}} & 
\T{\includegraphics[width=\imw]     {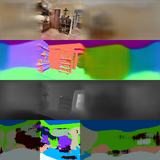}} & 
\T{\includegraphics[width=\imw]    {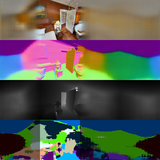}} & 
\T{\includegraphics[width=\imw]       {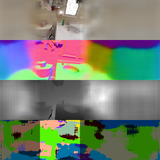}} \\
[+21pt]

\rotatebox[origin=c]{90}{G.T. 2} &
\T{\includegraphics[width=\imw]     {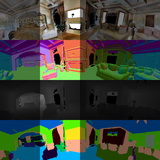}} & 
\T{\includegraphics[width=\imw]   {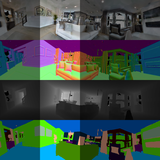}} & 
\T{\includegraphics[width=\imw]      {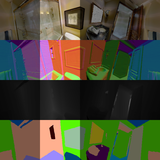}} & 
\T{\includegraphics[width=\imw]    {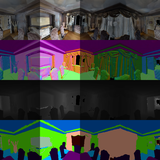}} & 
\T{\includegraphics[width=\imw]      {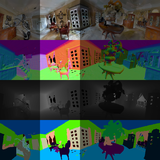}} & 
\T{\includegraphics[width=\imw]     {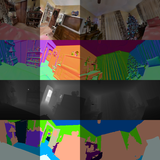}} & 
\T{\includegraphics[width=\imw]    {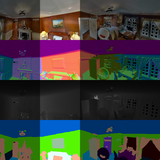}} & 
\T{\includegraphics[width=\imw]       {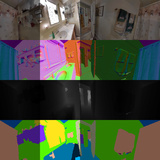}} \\
[+21pt]

\rotatebox[origin=c]{90}{Completed 2} &
\T{\includegraphics[width=\imw]     {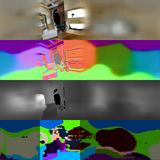}} & 
\T{\includegraphics[width=\imw]   {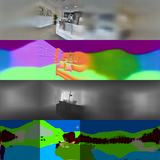}} & 
\T{\includegraphics[width=\imw]      {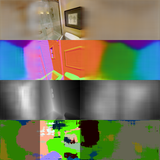}} & 
\T{\includegraphics[width=\imw]    {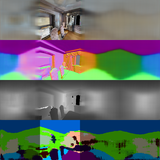}} & 
\T{\includegraphics[width=\imw]      {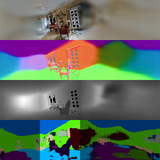}} & 
\T{\includegraphics[width=\imw]     {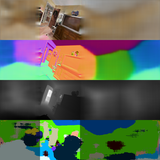}} & 
\T{\includegraphics[width=\imw]    {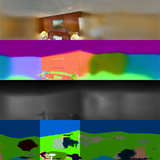}} & 
\T{\includegraphics[width=\imw]       {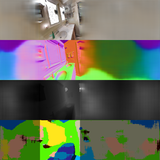}} \\ [-3pt]
\end{tabular}

\captionof{figure}{\small{Matterport qualitative results. From top to bottom: ground-truth color and scene geometry, our pose estimation results (two input scans in red and green), baseline results (4PCS, DL, GReg and CGReg), ground-truth scene RGBDNS and completed scene RGBDNS for two input scans. The unobserved regions are dimmed.}}
\label{Figure:Visualizations:Results:Matterport}
\end{figure*}

\begin{figure*}
\centering
\footnotesize
\def\imh{0.073\textwidth}
\def\imw{0.12\textwidth}
\newcommand{\T}[1]{\raisebox{-0.5\height}{#1}}
\setlength{\tabcolsep}{1pt}
\begin{tabular}{ccccccccc}

\rotatebox[origin=c]{90}{G.T. Color} &
\T{\includegraphics[width=\imw]     {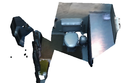}} & 
\T{\includegraphics[width=\imw]   {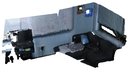}} & 
\T{\includegraphics[width=\imw]      {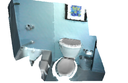}} & 
\T{\includegraphics[width=\imw]    {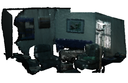}} & 
\T{\includegraphics[width=\imw]      {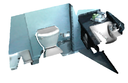}} & 
\T{\includegraphics[width=\imw]     {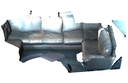}} & 
\T{\includegraphics[width=\imw]    {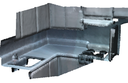}} & 
\T{\includegraphics[width=\imw]       {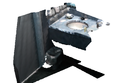}} \\
[+21pt]

\rotatebox[origin=c]{90}{G.T. Scene} &
\T{\includegraphics[width=\imw]     {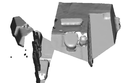}} & 
\T{\includegraphics[width=\imw]   {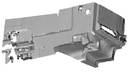}} & 
\T{\includegraphics[width=\imw]      {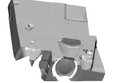}} & 
\T{\includegraphics[width=\imw]    {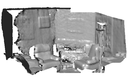}} & 
\T{\includegraphics[width=\imw]      {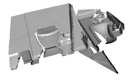}} & 
\T{\includegraphics[width=\imw]     {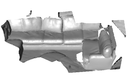}} & 
\T{\includegraphics[width=\imw]    {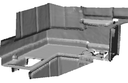}} & 
\T{\includegraphics[width=\imw]       {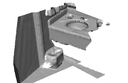}} \\
[+21pt]

\rotatebox[origin=c]{90}{Ours} &
\T{\includegraphics[width=\imw]     {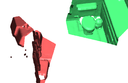}} & 
\T{\includegraphics[width=\imw]   {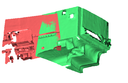}} & 
\T{\includegraphics[width=\imw]      {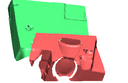}} & 
\T{\includegraphics[width=\imw]    {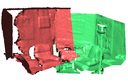}} & 
\T{\includegraphics[width=\imw]      {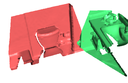}} & 
\T{\includegraphics[width=\imw]     {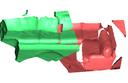}} & 
\T{\includegraphics[width=\imw]    {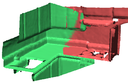}} & 
\T{\includegraphics[width=\imw]       {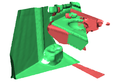}} \\
[+21pt]

\rotatebox[origin=c]{90}{4PCS} &
\T{\includegraphics[width=\imw]     {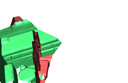}} & 
\T{\includegraphics[width=\imw]   {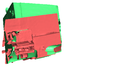}} & 
\T{\includegraphics[width=\imw]      {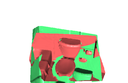}} & 
\T{\includegraphics[width=\imw]    {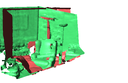}} & 
\T{\includegraphics[width=\imw]      {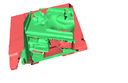}} & 
\T{\includegraphics[width=\imw]     {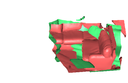}} & 
\T{\includegraphics[width=\imw]    {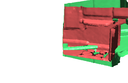}} & 
\T{\includegraphics[width=\imw]       {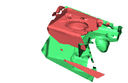}} \\
[+21pt]

\rotatebox[origin=c]{90}{DL} &
\T{\includegraphics[width=\imw]     {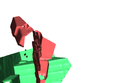}} & 
\T{\includegraphics[width=\imw]   {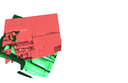}} & 
\T{\includegraphics[width=\imw]      {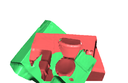}} & 
\T{\includegraphics[width=\imw]    {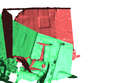}} & 
\T{\includegraphics[width=\imw]      {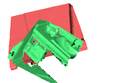}} & 
\T{\includegraphics[width=\imw]     {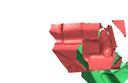}} & 
\T{\includegraphics[width=\imw]    {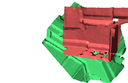}} & 
\T{\includegraphics[width=\imw]       {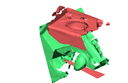}} \\
[+21pt]

\rotatebox[origin=c]{90}{GReg} &
\T{\includegraphics[width=\imw]     {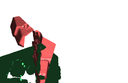}} & 
\T{\includegraphics[width=\imw]   {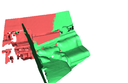}} & 
\T{\includegraphics[width=\imw]      {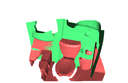}} & 
\T{\includegraphics[width=\imw]    {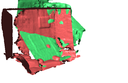}} & 
\T{\includegraphics[width=\imw]      {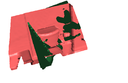}} & 
\T{\includegraphics[width=\imw]     {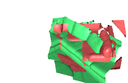}} & 
\T{\includegraphics[width=\imw]    {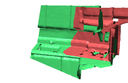}} & 
\T{\includegraphics[width=\imw]       {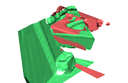}} \\
[+21pt]

\rotatebox[origin=c]{90}{CGReg} &
\T{\includegraphics[width=\imw]     {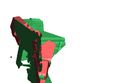}} & 
\T{\includegraphics[width=\imw]   {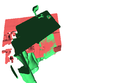}} & 
\T{\includegraphics[width=\imw]      {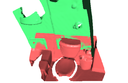}} & 
\T{\includegraphics[width=\imw]    {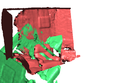}} & 
\T{\includegraphics[width=\imw]      {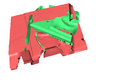}} & 
\T{\includegraphics[width=\imw]     {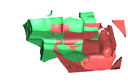}} & 
\T{\includegraphics[width=\imw]    {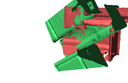}} & 
\T{\includegraphics[width=\imw]       {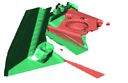}} \\
[+21pt]

\rotatebox[origin=c]{90}{G.T. 1} &
\T{\includegraphics[width=\imw]     {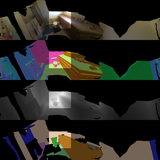}} & 
\T{\includegraphics[width=\imw]   {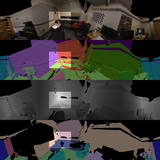}} & 
\T{\includegraphics[width=\imw]      {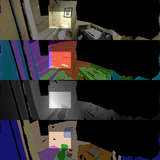}} & 
\T{\includegraphics[width=\imw]    {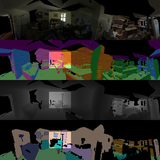}} & 
\T{\includegraphics[width=\imw]      {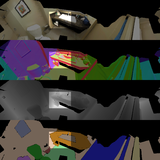}} & 
\T{\includegraphics[width=\imw]     {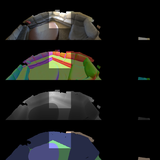}} & 
\T{\includegraphics[width=\imw]    {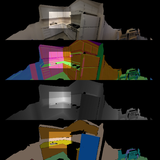}} & 
\T{\includegraphics[width=\imw]       {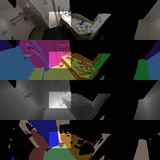}} \\
[+21pt]

\rotatebox[origin=c]{90}{Completed 1} &
\T{\includegraphics[width=\imw]     {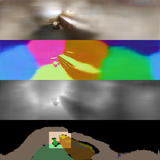}} & 
\T{\includegraphics[width=\imw]   {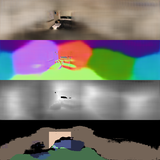}} & 
\T{\includegraphics[width=\imw]      {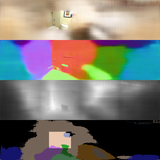}} & 
\T{\includegraphics[width=\imw]    {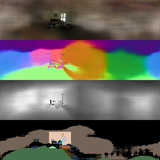}} & 
\T{\includegraphics[width=\imw]      {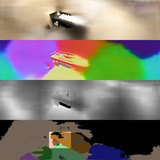}} & 
\T{\includegraphics[width=\imw]     {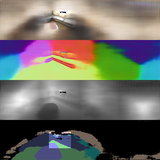}} & 
\T{\includegraphics[width=\imw]    {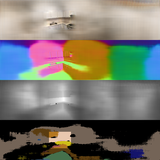}} & 
\T{\includegraphics[width=\imw]       {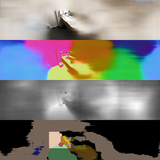}} \\
[+21pt]

\rotatebox[origin=c]{90}{G.T. 2} &
\T{\includegraphics[width=\imw]     {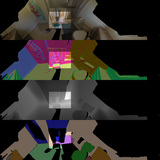}} & 
\T{\includegraphics[width=\imw]   {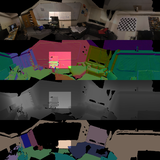}} & 
\T{\includegraphics[width=\imw]      {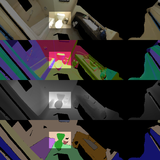}} & 
\T{\includegraphics[width=\imw]    {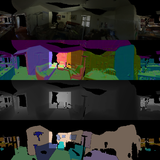}} & 
\T{\includegraphics[width=\imw]      {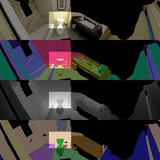}} & 
\T{\includegraphics[width=\imw]     {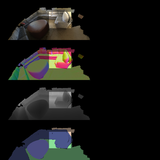}} & 
\T{\includegraphics[width=\imw]    {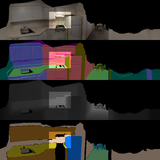}} & 
\T{\includegraphics[width=\imw]       {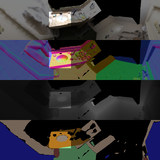}} \\
[+21pt]

\rotatebox[origin=c]{90}{Completed 2} &
\T{\includegraphics[width=\imw]     {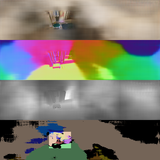}} & 
\T{\includegraphics[width=\imw]   {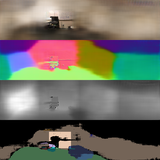}} & 
\T{\includegraphics[width=\imw]      {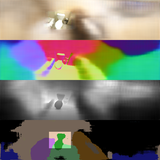}} & 
\T{\includegraphics[width=\imw]    {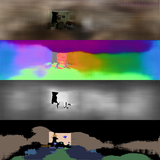}} & 
\T{\includegraphics[width=\imw]      {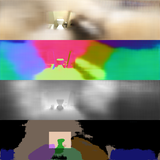}} & 
\T{\includegraphics[width=\imw]     {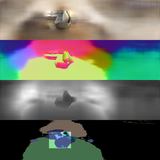}} & 
\T{\includegraphics[width=\imw]    {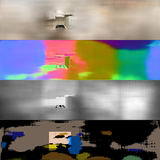}} & 
\T{\includegraphics[width=\imw]       {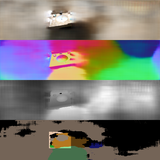}} 
\\ [-3pt]

\end{tabular}

\captionof{figure}{\small{ScanNet qualitative results. From top to bottom: ground-truth color and scene geometry, our pose estimation results (two input scans in red and green), baseline results (4PCS, DL, GReg and CGReg), ground-truth scene RGBDNS and completed scene RGBDNS for two input scans. The unobserved regions are dimmed.}}
\label{Figure:Visualizations:Results:ScanNet}
\end{figure*}

\end{document}